\documentclass[twoside,11pt]{article}

\usepackage{array} \newcolumntype{L}[1]{>{\raggedright\let\newline\\\arraybackslash\hspace{0pt}}m{#1}} \newcolumntype{C}[1]{>{\centering\let\newline\\\arraybackslash\hspace{0pt}}m{#1}} \newcolumntype{R}[1]{>{\raggedleft\let\newline\\\arraybackslash\hspace{0pt}}m{#1}}

\usepackage{nameref}
\usepackage{dnd}
\usepackage{times}
\usepackage{enumitem}
\usepackage{tipa}
\usepackage{tabularx}
\usepackage[dvipsnames]{xcolor}
\usepackage{fixltx2e}
\usepackage{float}
\usepackage{adjustbox}
\usepackage{amsmath}

\newenvironment{absolutelynopagebreak} {\par\nobreak\vfil\penalty0\vfilneg \vtop\bgroup} {\par\xdef\tpd{\the\prevdepth}\egroup \prevdepth=\tpd}

\newcounter{fpcounter} 

\renewcommand{\thefpcounter}{\arabic{fpcounter}} \newenvironment{fp}[2]{
\begin{absolutelynopagebreak} 
  \refstepcounter{fpcounter} 
  \label{#1} 
  \vspace{1em} \noindent{Example~\thefpcounter} \vspace{-0.9em}
\end{absolutelynopagebreak}
}
{}

\newcommand{\hold}{{\fontfamily{cmtt}\selectfont hold}}
\newcommand{\change}{{\fontfamily{cmtt}\selectfont change}}
\newcommand{\ih}{{\fontfamily{cmtt}\selectfont incomplete-hold}}
\newcommand{\qu}{{\fontfamily{cmtt}\selectfont question}}
\newcommand{\trailoff}{{\fontfamily{cmtt}\selectfont trail-off}}
\newcommand{\selfint}{{\fontfamily{cmtt}\selectfont self-interruption}}
\newcommand{\hrt}{{\fontfamily{cmtt}\selectfont hrt}}
\newcommand{\uncertain}{{\fontfamily{cmtt}\selectfont @}}
\newcommand{\combined}{{\fontfamily{cmtt}\selectfont label\_label}}

\newcommand{\cont}{{\fontfamily{cmtt}\selectfont cont}}
\newcommand{\particle}{{\fontfamily{cmtt}\selectfont part}}
\newcommand{\qupart}{{\fontfamily{cmtt}\selectfont q-part}}
\newcommand{\hes}{{\fontfamily{cmtt}\selectfont hes}}
\newcommand{\coll}{{\fontfamily{cmtt}\selectfont coll}}
\newcommand{\disruption}{{\fontfamily{cmtt}\selectfont disruption}}
\newcommand{\incomplete}{{\fontfamily{cmtt}\selectfont incomplete}}

\ShortHeadings{Turn-taking annotations}{Kelterer and Schuppler}
\firstpageno{1}

\begin{document}

\title{Turn-taking annotation for quantitative and qualitative analyses of conversation}

\author{\name Anneliese Kelterer \email anneliese.kelterer@uni-graz.at \\
       \addr Department of Linguistics\\
       University of Graz, Austria
       \AND
       \name Barbara Schuppler \email b.schuppler@tugraz.at \\
       \addr Signal Processing and Speech Communication Laboratory \\
       Graz University of Technology, Austria}

\maketitle

\begin{abstract}%
This paper has two goals. First, we present the  turn-taking annotation layers created for 95 minutes of conversational speech of the Graz Corpus of Read and Spontaneous Speech (GRASS),  available to the scientific community. Second, we describe the annotation system and the annotation process in more detail, so other researchers may use it for their own conversational data.
The annotation system was developed with an interdisciplinary application in mind. It should be based on sequential criteria according to Conversation Analysis, suitable for subsequent phonetic analysis, thus time-aligned annotations were made Praat, and it should be suitable for automatic classification, which required the continuous annotation of speech and a label inventory that is not too large and results in a high inter-rater agreement.
Turn-taking was annotated on two layers, Inter-Pausal Units (IPU) and points of potential completion (PCOMP; similar to transition relevance places).
We provide a detailed description of the annotation process and of segmentation and labelling criteria.
A detailed analysis of inter-rater agreement and common confusions shows that agreement for IPU annotation is near-perfect, that agreement for PCOMP annotations is substantial, and that disagreements often are either partial or can be explained by a different analysis of a sequence which also has merit. The annotation system can be applied to a variety of conversational data for linguistic studies and technological applications, and we hope that the annotations, as well as the annotation system will contribute to a stronger cross-fertilization between these disciplines.
 
\end{abstract}


\section{Introduction}
\label{sec:intro}

In the last two decades there has been an increasing interest in investigating speech phenomena of spontaneous conversations, both in speech science and speech technology \citep[cf., e.g.,][]{arnold2017words, Skantze2021, lopez2022evaluation}. In speech technology, the motivations to develop better models for spontaneous speech are, among others, to make automatic speech recognition more robust to variation, to make speech synthesis sound more natural and to make speaking robots more social. In speech science and linguistics, the aim is to reveal mechanisms of every-day speech processing that would not show in laboratory conditions (e.g., see \cite{TuckerErnestus2016} on why psycholinguistics needs to investigate casual speech, and \cite{Wagner2015} on why stylistic diversity is relevant to speech research in general). In the view of \citet[p. 10]{Wagner2015}, “phonetics does not need to continue the debate about what is the most appropriate data, but we need a better understanding about how our methods, including the speaking style recorded, influence our results.” 
Studying data from spontaneous conversations comes with several methodological challenges. Given the higher degree of variation than in less spontaneous speaking styles, more data needs to be available for analysis, and more complex statistical and acoustic analysis methods need to be developed \citep[e.g.,][]{Ward2019, ludusan2022analysis, WU202345}. From the common need for data and annotations for conversational speech, there has emerged a strong cross-fertilization between quantitative phonetics and speech technology (for a survey see \citep{SCHUPPLER2024103007}). Given that variation at different levels (e.g., pronunciation, lexical choice, syntactic structure and turn-taking) may interact and their separate quantitative analysis may thus not allow a holistic interpretation, qualitative analysis methods are also fundamental to increase our understanding of everyday speech processing. The qualitative paradigm relevant for this paper is Conversation Analysis \citep[CA; e.g.,][]{Sacks1974}. With respect to qualitative methods, there has been hardly any cross-fertilization with the speech technology community, and its usage in technology has been limited to error analysis, i.e., analyzing those tokens where model predictions were incorrect. The present paper presents a turn-taking annotation system for quantitative and qualitative analyses of conversations, useful for linguists, speech scientists and speech technologists\footnote{We presented a previous state of the annotations in GRASS, including a short description of a preliminary version of the annotation system, at the \textit{First Workshop on Integrating Perspectives on Discourse Annotation} \citep[\textit{DiscAnn}; subsequently published as ][]{ schuppler2021developing}. Since then, the annotation guidelines have been revised, more data has been annotated and previous annotations corrected. The present paper is moreover different from the presentation at the workshop in that it includes a more extensive description of the theoretical background, and of segmentation and labelling criteria. The present paper also includes a detailed description and evaluation of the annotation process, as well as possible applications, which were not included in the presentation at \textit{DiscAnn}.}.

The goals of this paper are twofold. The first goal is to present an extension to the \textit{Graz corpus of Read And Spontaneous Speech} \citep[GRASS;][]{schuppler2017corpus}. GRASS was originally published with orthographic annotations only. We annotated a subset of the recordings with turn-taking annotations on two layers; one based on the domain of Inter-Pausal Units (IPU) and one based on the CA concept of potential turn completion \citep[PCOMP; cf., transition relevance place;][]{Sacks1974}. 
Turn-taking annotations on these two layers have been the basis of investigations of various aspects in different disciplines. For instance, the prosody of turn-taking in the domain of Inter-Pausal Units was investigated by \cite{wichmann-caspers-2001-melodic}, \cite{EdlundHeldner2005}, \cite{gravanohirschberg2011}, \cite{Hjalmarsson2011}, \cite{Levitan2015}, and \cite{Brusco2020}. On the basis of similar operationalizations of transition relevance places\footnote{While these points are based on the concept of transition relevance places, these authors did not include prosodic criteria in their definition, since prosody was the object of investigation in these studies. We followed this approach; see also definition of PCOMP in Section \ref{sec:PCOMPdomain}.}, prosody was investigated by \cite{LocalWalker2012}, \cite{Zellers2017},  \cite{zellers2019hand}, and \cite{Enomoto2020}. The turn-taking annotations in GRASS provide a basis for similar, as well as a variety of other, investigations in Austrian German (cf., Sections \ref{sec:dynamics_PCOMP} and \ref{sec:conclusion}).

The second goal of this paper is to present the annotation system itself along with a detailed report of the annotation process. 
This detailed report is of interest to  those scientists who plan to apply the annotation system to their own conversational speech data. Even though the annotated corpus contains casual conversations between healthy speakers of Austrian German, the system is designed such that it is language independent, and it may be applied to healthy or pathological speech of speakers of any age. Another characteristic of the present annotation system is that it is suitable both for linguistic studies as well as for speech technology based applications, resulting from the fact that it was developed within a project that brought together linguists and speech technologists to study the prosody of turn-taking and to integrate this information into language models for conversational speech \citep{schuppler2021developing}. The requirements to the annotation system thus were given by the different disciplines' perspectives. The annotation system should be suitable for:

    \begin{enumerate}[label=(\roman*)]
        \item investigations informed by Conversation Analysis: the identity of a label is based on what happens next in the conversation, thereby avoiding interpretation (“it sounded like the speaker actually wanted to continue speaking, but…”).
        \item quantitative phonetic analysis: annotations are made in Praat \citep{praat} in interval-tiers in order to have time-stamps to all turn-taking units, allowing the extraction of timing and acoustic features, and relating the turn-taking annotations to other layers of annotation (orthography, phone segmentation, prosody).
        \item automatic classification of the turn-taking labels: in order to build a reasonably good classifier, we need a continuous stream of speech annotated in order to be able to make use of context information. Furthermore, the label inventory should not be too large given the amount of data in total and should be so concrete and clear that it  results in a high inter-rater agreement.  
    \end{enumerate}

Taking these considerations into account, we created turn-taking annotations on two separate interval tiers in Praat: a) speech segmented into Inter-Pausal Units (IPU), and b) speech segmented into points of potential completion (PCOMP) based on syntactic-pragmatic criteria, but not on phonetic/prosodic criteria to avoid circularity \citep[cf.,][]{LocalWalker2012, Zellers2017}. Section \ref{sec:TT_annotations} provides more detailed descriptions of these two annotation layers. Overall, the annotation system has a wide application range, both with respect to the type of the speech investigated and with respect to different quantitative and qualitative approaches.

The next section presents considerations for the development of our turn-taking annotation system. In Section \ref{sec:GRASS}, we present GRASS, giving an overview over the recording setting, speaker characteristics and speech style, and over other existing annotations for this corpus. In Section \ref{sec:TT_annotations}, we present the two turn-taking annotation layers, including the set of labels and their distribution in the annotated data. Moreover, we present more detailed annotation guidelines aimed at researchers who wish to apply this annotation system to their own corpus, with examples for segmentation and labeling on the two layers, and a validation of the annotations, including a detailed analysis of disagreements between annotators. In Section \ref{sec:dynamics}, we present a pilot study of conversational dynamics that can be captured with these annotations, and how these annotations can inspire new research questions. Section \ref{sec:conclusion} concludes the paper and lists further possible research questions that could be answered with these annotations.

\section{Considerations for the development of the turn-taking annotation system}
\label{sec:considerations}

One requirement for the annotation system was that it should be comprehensive (cf., III. above), that is, it should be possible to annotate all utterances that occur continuously in a conversation, not just a selection of pre-defined behaviours \citep[as is best practice within CA, e.g.,][]{StiversEnfield2010}. One annotation system that meets this requirement is SWBD-DAMSL \citep{Jurafsky1998}. However, while this annotation system also includes some tags that refer to turn-taking, it mixes turn-taking criteria (Acknowledge (Backchannel), Hold before answer/agreement) with attitudinal criteria (Agree/Accept, Appreciation), syntactic criteria (Wh-Question, Statement-non-opinion) and other CA concepts \citep[Conventional-closing, Dispreferred answers, Apology; cf.,][p. 6 for a list of labels]{Jurafsky1998}. Another annotation system meeting this requirement is the one presented in \citeauthor{gravanohirschberg2011}'s \citeyearpar{gravanohirschberg2011} investigation of task-based dialogs. However, we wanted a more fine-grained annotation that is rooted in CA criteria, which are more appropriate for the conversational style used in GRASS. For instance, they did not distinguish between questions (other selection) vs. other kinds of turn changes (self-selection). A third annotation system meeting this requirement is the one developed by \cite{Enomoto2020}. This system is based on CA criteria \citep{Sacks1974}, but it was published only after we had already trained the annotators and they had started annotating GRASS. Their annotation system differs from ours in two main aspects. First, they use forward-looking as well as backward-looking labels, while our labels are primarily forward looking (cf., Section \ref{sec:TT_annotations}). Second, they mark overlap in their labels, while overlap is retrievable from the time-aligned annotations in our system, which allows for a precise analysis of timing, and for a more fine-grained, time-sensitive investigation of different kinds of overlap (e.g., with respect to reaction times).

One annotation system that seemed particularly suitable for requirements I. and II. above is the one used by \cite{Zellers2017}. Zellers distinguished between \hold, when the current speaker continued speaking; \change, when a speaker change occurred and another speaker continued speaking; \qu, when a current speaker actively transferred the turn to another speaker by asking a question; and {\fontfamily{cmtt}\selectfont backchannel}, when a speaker produced a short hearer response token \citep[cf., terminology in][]{Sikveland2012} that did not take up the turn. We adopted this system by \cite{Zellers2017}. To be able to characterize all of interlocutors’ speech production in longer sequences (cf., III. above), however, we required an extension of the label set. 
For instance, the annotation system should be able to distinguish syntactically dependent from independent turn-continuations, to distinguish between syntactically complete and incomplete turn-holds, 
to capture syntactically incomplete turn-holds with subsequent rephrasing, to represent different kinds of discourse and hesitation particles that are independent of syntax, and to characterize different kinds of syntactically incomplete turn-changes. 
This mixed bottom-up approach resulted in seven labels on the IPU layer and in eleven labels on the PCOMP layer (cf., Sections \ref{sec:IPUlabels} and \ref{sec:PCOMPlabels}).
We chose a separation into two different layers for several reasons. A variable of an annotation system should “not conflate properties that are potentially independent of each other” \cite[p. 124]{Stefanowitsch2020}. Pausing and transition relevance places (TRP) are independent of each other in speech. Several TRPs can occur before the next pause, and several pauses can occur before a speaker reaches the next TRP. Separating the annotation into two layers made it easier for annotators, because they did not have to consider two independent phenomena at the same time when segmenting and choosing labels. Moreover, the separation into IPU and PCOMP reduced the set of labels, which is also easier to annotate, and easier to classify. Questions about the relationship between pausing and syntactic completion can then be addressed by combining information from both layers (e.g., where do boundaries on the IPU layer coincide with boundaries on the PCOMP layer, and where are IPU boundaries not accompanied by a PCOMP boundary). 

\newpage
\section{GRASS corpus}
\label{sec:GRASS}

The Graz corpus of Read and Spontaneous Speech \citep[GRASS;][]{schuppler2017corpus} is a corpus containing 4.6 hours of read speech and 19 hours of conversational speech from 38 speakers of Austrian German. The read speech component of this corpus is not relevant for this paper and are described in \cite{schuppler2017corpus}. The conversational component consists of 19 one hour-long dyadic conversations recorded in a recording studio. Speakers were recorded with head-mounted microphones on separate channels. Nevertheless, in overlapping speech, the speech of the interlocutor is still often audible on the current speaker's channel. The speaker sample is balanced for gender (6 f-f, 7 f-m, 6 m-m) and comprises 20-60 year old native speakers of Austrian German from the South Bavarian and southern Middle Bavarian dialect area from the south-east of Austria. Speakers were originally from the federal states of Styria, Carinthia, Salzburg, Upper Austria and Burgenland and had lived in Graz for most of their adult life. The speakers in each conversation knew each other well and were friends, partners, family members or colleagues. No task was given for the conversations and no experimenter was present during the recordings, resulting in a casual speech style. \cite{schuppler2017corpus} reported the exact recording procedure and setting, as well as a description of features characterising the speech style. Speakers use a mix between Standard Austrian German and South Bavarian or southern Middle Bavarian dialects. While individual speakers occupy different positions along the Standard-dialect continuum, there is also considerable within-speaker variation along this parameter, with all speakers using both Standard Austrian German and dialectal forms, albeit to different degrees \citep{geiger23_interspeech}.

For these recordings, time-aligned annotations are available in Praat for a basic orthographic transcription \citep{schuppler2017corpus}, automatic phonetic segmentation alignments \citep{linke2023} and prosody \citep{schuppler2017corpus}. The basic annotation layer of GRASS consists of a comprehensive orthographic transcription of all vocal productions (speech and non-speech noises) in the 19 conversations. This layer also marks overlapping speech, disfluencies, and non-speech noises (coughing, smacks, breathing (in vs. out) and laughter). Words were annotated with standard German orthography rather than a graphemic representation of phonetic realizations to facilitate search-ability and comparability of various forms of the same word. 

Word and phone-level segmentations and labels were created by means of a Kaldi-based forced-alignment tool for the whole of GRASS \citep{linke2023}. The tool specifically made use of pronunciation variants to capture various possible pronunciations along the standard-dialect continuum (e.g., [a\textipa{:}] $\sim$ [\textipa{A:}] $\sim$ [\textipa{0:}] or /l/ vocalisation) as well as on the canonical-reduced continuum \citep[cf.;][]{schuppler2014pronunciation}.

For a sub-portion of the conversational speech component of GRASS (115 min in 19 conversations) manually created prosodic annotations are available \citep{schuppler2017corpus}. The prosodic annotations consist of annotations for prosodic phrase boundaries, word-level (perceptual) prominence and 3) pitch events. The data annotated on the IPU and PCOMP level (i.e., the concrete stretches of speech from the conversations) presented in Sections \ref{sec:IPU} and \ref{sec:PCOMP} are a subset of the data annotated at the prosodic level. This choice was made to facilitate analyses of prosodic characteristics of turn-taking.
\clearpage

\section{Turn-taking annotations in GRASS}
\label{sec:TT_annotations}

This section presents the turn-taking annotations that expand the annotation of GRASS (including a presentation of the criteria of segmentation and labelling; cf., Section \ref{sec:GRASS}), a description of the annotation process, and a detailed validation of these annotations. See Section \ref{sec:availability} for the availability of this corpus to the scientific community.

\begin{figure}[ht]
    \centering
    \includegraphics[width=\linewidth]{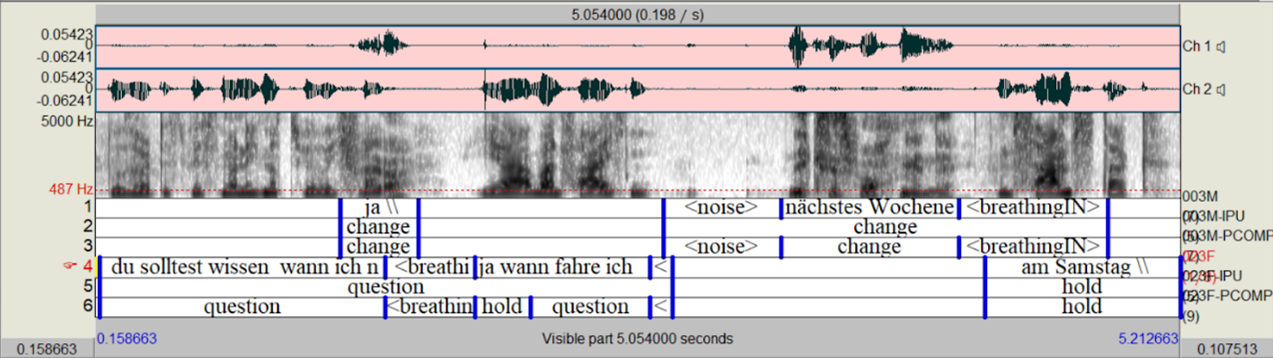}
    \caption{Praat screenshot of orthographic, IPU and PCOMP annotations for two speakers (003M: first three rows; 023F: last three rows), illustrating the temporal organization of questions, turn-holds and turn-changes.}
    \label{fig:Praat}
\end{figure}

Annotation (segmentation and labelling) was done in Praat, that is, in a time-aligned manner in a program that allows for exact boundaries according to phonetic criteria (e.g., between two adjacent words). Timing, such as pauses or overlapping speech is not annotated explicitly (as is usual in CA transcription schemes), as this information can be extracted from the time-aligned annotations. Figure \ref{fig:Praat} shows an example. The first three annotation rows refer to the speaker recorded in channel 1 (003M\footnote{All examples including speaker IDs in the form of 003M, 023F, 005M, 028F, etc., refer to examples from GRASS. A list of all examples from GRASS with their respective time stamps in the corpus is presented in Table \ref{tab:timestamps} in \nameref{sec:appendixA}. Invented examples are marked with speaker A and speaker B}) and the last three rows refer to the speaker in channel 2 (023F). For each speaker, the first row presents the orthographic transcription, the second row the IPU, and the third the PCOMP annotation. Breathing and other non-speech articulation noises are annotated separately on the orthographic and on the PCOMP tier, but are included in IPU intervals (cf., \ref{sec:IPUdomain}). 
This example shows that the answer to 023F's first question (“ja” - 'yes') by speaker 003M (labelled as change on both IPU and PCOMP) overlaps partly with the end of the question (“du solltest wissen wann ich nach Zürich fahre?” - ‘you should know when I’m going to Zurich?’)\footnote{This question is segmented and labelled as \qu \hspace{0.5mm} on PCOMP, but integrated into a larger questioning IPU, cf., segmentation criteria for IPU and PCOMP in Sections \ref{sec:IPUdomain} and \ref{sec:PCOMPdomain}} and partly with the following inbreath. This example also shows that 003M only answers 023F's second question (“ja? wann fahre ich denn?” - ‘yeah? When am I going?’) after a pause that is filled by a glottal articulation (labelled as $<$noise$>$). After 003M’s answer (“nächstes Wochenende” - ‘next weekend’), but in overlap with the subsequent inbreath, 023F responds with “am Samstag” (‘this Saturday”).

\subsection{Inter-Pausal Units (IPU)}
\label{sec:IPU}

\subsubsection{Domain of IPU annotations}
\label{sec:IPUdomain}

The first layer of turn-taking annotations is in the domain of Inter-Pausal Units (IPU). These units have been defined with different pause duration thresholds in the literature \citep[e.g.,][]{gravanohirschberg2011,bigi:hal-02428485}. 
We defined an IPU as a stretch of speech, including audible breathing\footnote{Since there are no clear guidelines about which breathings should be considered as interactionally functional (such as sharp inbreaths to indicate the beginning of a turn) and which should not, we included all audible breathings in IPUs.} and smacks, that is separated from other speech produced by the same speaker by a silence of at least 150 ms. We chose this threshold because it is longer than most voiceless plosive durations and facilitates automated IPU boundary detection in the future (i.e., minimizing the misdetection of plosive durations as pause durations). Even though we had future automatic segmentation in mind, IPU segmentation was done manually by the same annotators that did the labelling (cf., Section \ref{sec:IPUprocess}).

There is only one exception to the IPU segmentation guideline of a pause of $>$ 150 ms, which is when a long IPU ends in a hearer response token (\hrt, see Section \ref{sec:IPUlabels}; usually separated from previous speech by an audible breathing or other kind of mouth noise). In this case, the hearer response token is segmented separately, even though it is not separated from previous speech by a silent pause. We introduced this exception because of the predominantly forward-looking nature of the IPU labelling scheme, so that a long IPU including other functions is not labelled as \hrt. 

IPU labels can be conceptualised relatively independently of the domain. If our segmentation criterion had been different (e.g., counting audible breathing as a pause, or automatic annotation based on voice activity detection), the labelling criteria would still be the same. If, for instance, IPU labels were applied to IPUs separated by pauses $>$ 50 ms, there would be a larger number of IPUs, but whether a specific IPU was labelled as e.g., \hold  \hspace{0.5mm} or \change, would still be decided based on the same criteria. Therefore, the IPU annotation layer is suitable for coarser turn-taking annotation and is applicable to automatically segmented IPUs, even if criteria for automatic IPU segmentation differs from our approach here.

\subsubsection{IPU labels}
\label{sec:IPUlabels}

\begin{figure}
\caption{Decision tree for assigning single IPU labels. A decision tree for assigning IPU single as well as combined labels is presented in Figure \ref{fig:Flowchart_IPUcombined} in \nameref{sec:appendixB}.}
\centering
\includegraphics[scale=0.7]{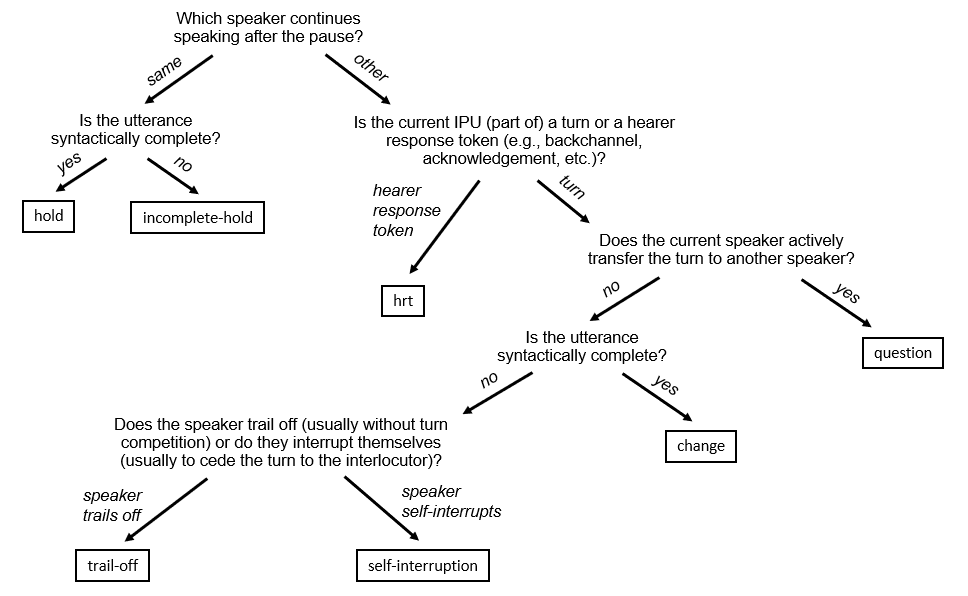}
\label{fig:Flowchart_IPUsimple}
\end{figure}

IPU labels can be grouped into three categories: A) the same speaker continues with their turn after the pause, B) a turn-transition occurs, that is, the other speaker continues talking after, or often already in overlap, with the current IPU, and C) speech productions that do not constitute a turn of their own and that do not interrupt the interlocutor’s turn. Table \ref{tab:IPUlabels} presents a list of the IPU labels with short definitions, and Figure \ref{fig:Flowchart_IPUsimple} presents a decision tree for assigning single IPU labels\footnote{A decision tree for assigning single as well as combined IPU labels (cf., Section \ref{sec:IPUcombined}) is presented in Figure \ref{fig:Flowchart_IPUcombined} in \nameref{sec:appendixB}.}. 

Category A) contains two labels:

\begin{itemize}
  \item \hold, when a speaker produces a pause at a transition relevance place (TRP; i.e., a point of syntactic/pragmatic completion, not taking into account prosodic completion) and continues speaking after the pause. Since these labels are all forward-looking, the beginning of a turn is not marked by these labels. That is, a turn-initial and a turn-medial hold are both labelled as \hold.
  \item \ih, when a speaker produces a pause not at a TRP, but at a point of maximum grammatical control \citep{schegloff1996turn}, and continues speaking after the pause. Typical {\ih}s end in “but”, “and then”, “the”, etc., after which a pause is produced. In many cases, the structure is continued after the pause, e.g. (“vielleicht soll es einfach nur… gut klingen manchmal” (‘maybe it’s just supposed to… sound cool sometimes’, 025F). However, speakers might also start to rephrase themselves after the pause. The essential criterion for the \ih \hspace{0.5mm} label is that the speaker pauses in the middle of a syntactic structure and that the same speaker continues talking after the pause.
\end{itemize}

Category B) contains four labels: 

\begin{itemize}
  \item \change, when an IPU occurs before a speaker change when the IPU ends in a TRP. Longer turns in which a speaker produces several turn-internal pauses consist of several IPUs in which pre-final IPUs are labelled as \hold{} or \ih{} and the last IPU before the speaker change is labelled as \change. Shorter turns might consist of only one short IPU labelled as \change.
  \item \qu, when a current speaker actively transfers the turn to the next speaker. This is often done in the form of a syntactic question, but can also take the form of a prosodic question or suggestion that the interlocutor knows something of interest (e.g., “du weißt das sicher” – ‘surely you know this’/‘you know this, don’t you’). Crucially, only functional questions, that is, those that are treated as such and are answered, are labelled as \qu, but not, for instance, self-directed questions (“what was I saying?”) or rhetorical questions. If an interlocutor interprets a rhetorical question as an actual question, however, and provides an answer, the question is labelled according to the interlocutors’ behaviour and not according to speaker intention, that is, as \qu. If an interlocutor does not answer immediately, the first speaker sometimes rephrases the question (e.g. 005M: ist das irgendwie sprachspezifisch oder… ist das wurscht? … an was für einem Spracherwerb die jetzt forschen?” – ‘is it somehow language specific or … doesn’t it matter? … what kind of language acquisition they are researching?’) or adds some specification (e.g., “did you have fun? … at the concert yesterday?”), resulting in a series of questions. In this case, only the last IPU that actually transfers the turn to the interlocutor is labelled as \qu. This issue is resolved by a different operationalisation of questions on the PCOMP layer, which can capture that the previous IPUs, or previous utterances in the same IPU were also (syntactic/prosodic/pragmatic) questions (cf., Example \ref{fp:wohnen001002} in Section \ref{sec:PCOMPcombined}).
  Seeking confirmation, even when done with e.g., interrogative prosody (e.g., “really?”) is not treated as a question in this annotation system but as a continuer (see below). Repair questions (e.g., “hm?” when there was a problem of understanding), however, are labelled as \qu.

  \item \trailoff, when a speaker produces a pause in the middle of a syntactic structure, at a place of maximum grammatical control, after which the interlocutor takes up the turn, often after quite a long pause. This includes trail-off conjunctions as described by \cite{walker2012trailoff}, as well as speech productions that might have been designed as turn-holding, but in which the interlocutor took up the turn in the following pause.
  \item \selfint, when speakers interrupt themselves, often in the middle of a prosodic word. This often happens in turn competition, when both interlocutors start speaking at the same time, or when one interlocutor starts speaking during the other one’s turn and one of them interrupts themselves to cede the turn to the other. 

\end{itemize}

Category C) only contains one label for various kinds of hearer response tokens;

\begin{itemize}
  \item \hrt. This label marks hearer response tokens, such as backchannels, 
  continuers, 
  acknowledgement tokens, 
  short agreements (e.g., “ja stimmt” – ‘yeah that’s right’) and tokens expressing sympathy (e.g., 'oh dear'). 
  Hearer response tokens in GRASS are usually very short (“mhm” and “ja” are most frequent), but they may also be longer, for instance, when an interlocutor repeats part of what the current speaker just said to indicate that they are listening, for instance, A: “Ich war gestern im Kino” – B: “ah, im Kino” (A: ‘I was at the cinema yesterday’ – B: ‘ah, at the cinema’). It is important to note that hearer response tokens do not take over the turn but may influence what is being said. For instance, “echt” (‘really’) with a rising intonation often elicits more talk on the current subject. 
  \hrt s do not interrupt another speaker’s turn. Thus, if an interlocutor produces a \hrt \hspace{0.5mm} in the current speaker’s pause, the current speaker’s previous IPU is labelled as \hold, even though the \hrt \hspace{0.5mm} is inserted between two IPUs, cf., Example \ref{fp:004M024Fspektatulär}. \hrt s are sometimes produced in alternation by both speakers. 
  
\setlength{\leftskip}{1cm}
\begin{fp}{fp:004M024Fspektatulär}{} 
\begin{center}
\includegraphics[width=\linewidth]{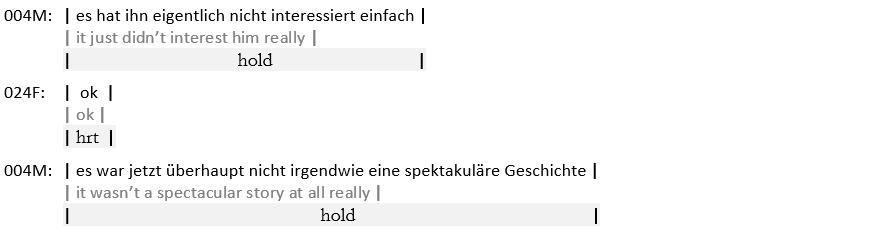}
\label{ex:004M024Fspektakulaer}
\end{center}
\end{fp}
\setlength{\leftskip}{0cm}
\vspace{-3em}

  Answers to questions, even if they consist only of “ja”, “mhm” or “ok” were not annotated as \hrt, but as \change{} or as \hold{} if the speaker continued after the answer.
  {\hrt}s might also be followed by more talk by the interlocutor producing the \hrt \hspace{0.5mm} instead of the other interlocutor. This happens, for instance, if a speaker produces an acknowledgement token and then takes up the turn, often changing the (sub)topic.

\end{itemize}

Apart from turn-taking criteria (same or other speaker continues, interlocutor produces a hearer response token), IPU labels can also be grouped into whether they end in a TRP (\change, \hold, \qu) or whether a pause is produced in the middle of a syntactic structure (\ih, \selfint, \trailoff). The relatively high percentage of \ih, compared to all turn-holds (39\%) shows that pausing at a syntactically incomplete point is a strategy that speakers employ frequently to hold their turn. In contrast, syntactically incomplete structures occurred less frequently at turn boundaries (17\% of all turn-changes were \selfint{} or \trailoff).

The distribution of single IPU labels in different conversations in GRASS is presented in Table \ref{tab:IPUcounts}. Overall, 2593 IPUs were annotated in 95 minutes in 15 conversations in GRASS (10 minutes in 4 conversations, 5 minutes in 11 conversations). 2292 IPUs were labelled with single labels (91.6\%), 44 IPUs with combined labels (1.8\%; cf., Section \ref{sec:IPUcombined}) and 167 IPUs were labelled as uncertain (6.7\%; cf., Section \ref{sec:IPUprocess}). 

\begin{table}[ht!]
\begin{center}
\caption{List of IPU labels and their definitions.}
\vspace{+5pt}

\begin{tabular}{p{0.22\textwidth}p{0.72\textwidth}}
  \hline
 \textbf{IPU label} & \textbf{Definition} \\  \hline
 \hold &  speaker makes a pause at a syntactically complete point and continues speaking\\  \hline
 \ih &  speaker makes a pause at a syntactically incomplete point and continues speaking\\  \hline
 \change & speaker makes a pause at a syntactically complete point and the other speaker continues speaking \\   \hline
 \qu & speaker actively transfers the turn to the other speaker with a functional question that is subsequently answered (excluding, e.g., rhetorical questions) \\   \hline
 \trailoff & speaker makes a pause at a syntactically incomplete point and the other speaker continues speaking \\   \hline
 \selfint & speaker makes a pause in the middle of a syntactic construction, often in the middle of articulation, often to cede the turn to the other speaker in overlap \\   \hline
 \hrt & hearer response token; usually short backchannels, continuers, acknowledgements, etc., that do not contain a (new) proposition of their own and don’t take up the turn \\   \hline
 \combined & combination of two of the labels above (in alphabetical order). Common cases are described in Section \ref{sec:IPUcombined} \\   \hline
 \uncertain & indicates uncertainty about a label (cf., Section \ref{sec:IPUprocess}); may also co-occur with a combined label to indicate the uncertainty between two specific labels (cf., Section \ref{sec:IPUvalidation}) \\       \hline
\end{tabular}
\label{tab:IPUlabels}
\end{center}
\end{table}

523 IPUs were labelled as \hold \hspace{0.5mm} (23\%), 334 as \ih \hspace{0.5mm} (15\%), 501 IPUs as \change \hspace{0.5mm} (22\%), and 178 as \qu \hspace{0.5mm} (8\%), while syntactically incomplete turn-changes occurred less frequently (\selfint: n = 74, 3.2\%; \trailoff: n = 43, 1.9\%). The most frequent type was \hrt, which was labelled in 641 IPUs (28\%). Table \ref{tab:IPUcounts} illustrates that some conversations are more balanced (e.g., 001M002M), while other conversations show a clear asymmetry. For instance, 015M produced more instances of \hold, \ih \hspace{0.5mm} and \change \hspace{0.5mm} while his interlocutor 017M produced more instances of \qu \hspace{0.5mm} and \hrt. How IPU annotations capture conversational dynamics is discussed in more detail in Section \ref{sec:dynamics_IPU}.

\begin{table}[ht!]
\centering
\caption{Distribution of single IPU annotations per speaker, sorted by conversation.}
\vspace{+5pt}
\begin{tabular}{p{0.05\textwidth}R{0.05\textwidth}R{0.15\textwidth}R{0.08\textwidth}R{0.12\textwidth}R{0.1\textwidth}R{0.15\textwidth}R{0.08\textwidth}}
  \hline
spkID & \small\hold & \small\ih & \small\change & \small\qu & \small\trailoff & \small\selfint & \small\hrt \\ 
  \hline
001M &  19 &   7 &  25 &   3 &  &   4 &  32 \\ 
  002M &  13 &   2 &  26 &   5 &  &  &  30 \\  \hline
  003M &  18 &  17 &  17 &   2 &   1 &   1 &   8 \\ 
  023F &  11 &   4 &   9 &  12 &  &  &   5 \\  \hline
  004M &  19 &   6 &  24 &   4 &  &   2 &   9 \\ 
  024F &  10 &  14 &  18 &   8 &   1 &   1 &  15 \\  \hline
  006M &  42 &  13 &  25 &   9 &   2 &  &  13 \\ 
  007M &  30 &   9 &  24 &  12 &  &   2 &  28 \\  \hline
  009M &  11 &   9 &  16 &   1 &   1 &   5 &   7 \\ 
  010M &   3 &  &  15 &   6 &  &   5 &  20 \\  \hline
  013M &  10 &   5 &   8 &   8 &   1 &   3 &  52 \\ 
  014M &  28 &   9 &  12 &   1 &  &   2 &  19 \\  \hline
  015M &  17 &   8 &  29 &  &   2 &   9 &   8 \\ 
  017M &   2 &   2 &   8 &  29 &   1 &   1 &  22 \\  \hline
  021F &   5 &   3 &   6 &   1 &   2 &  &  38 \\ 
  022F &   5 &   8 &   5 &   1 &  &   2 &  20 \\  \hline
  005M &  31 &  36 &  21 &  14 &   2 &  11 &  42 \\ 
  025F &  35 &  19 &  38 &   9 &   1 &   2 &  26 \\  \hline
  026F &   3 &   3 &   7 &   3 &   3 &  &  62 \\ 
  027F &  35 &  19 &   4 &   1 &   2 &   6 &  14 \\  \hline
  008M &  40 &  23 &  35 &  16 &   1 &   3 &   9 \\ 
  028F &  23 &  18 &  38 &   6 &  10 &   5 &  11 \\  \hline
  029F &   7 &   3 &  17 &   3 &   1 &  &  18 \\ 
  030F &  47 &  41 &  14 &   2 &   4 &   1 &   8 \\  \hline
  011M &   7 &  12 &  11 &   2 &   1 &  &   6 \\ 
  031F &  10 &   3 &  10 &   6 &   2 &   3 &  38 \\  \hline
  012M &  13 &   9 &  18 &   1 &   1 &  &  18 \\ 
  032F &  14 &  12 &   7 &   7 &  &   2 &   7 \\  \hline
  038F &   8 &  17 &  10 &  &   3 &   1 &   3 \\ 
  039F &   7 &   3 &   4 &   6 &   1 &   3 &  53 \\  \hline
  TOTAL & 523 & 334 & 501 & 178 &  43 &  74 & 641 \\ 
   \hline
\end{tabular}
\label{tab:IPUcounts}
\end{table}

\subsubsection{IPU combined labels}
\label{sec:IPUcombined}
To account for the full range of interlocutors’ behaviours, we allowed for the combination of the labels described in Section \ref{sec:IPUlabels}. We introduced these combined labels for when more than one label applied. 
Depending on the research question, these combined labels easily allow for their grouping with either one or the other label, for their exclusion from the data set, or for the investigation of complex speaker behaviour by specifically looking at these labels.
It is important to note that these combined labels indicate complex speaker behaviours in the communicative context, and not an annotator’s uncertainty. Uncertainties, on the other hand, were indicated by adding \uncertain \hspace{0.5mm} to a label or to a combination of labels if the uncertainty was between two specific labels (cf., Section \ref{sec:IPUprocess}).
\par
Overall, combined IPU labels occurred very rarely (N = 49; 1.8\%). 14 different label combinations were annotated, most of them only in one or two instances. Four combined IPU labels, however, occurred more frequently; {\fontfamily{cmtt}\selectfont change\_hold} (n = 9), {\fontfamily{cmtt}\selectfont change\_hrt} (n = 6), {\fontfamily{cmtt}\selectfont hold\_hrt} (n = 9) and {\fontfamily{cmtt}\selectfont question\_trail-off} (n = 4). Annotators labelled {\fontfamily{cmtt}\selectfont change\_hold} when both the same and the other speaker talked after the pause (i.e., before overlapping speech) and when the current IPU was followed by a short utterance expressing some attitude by the other speaker (e.g., incredulity, indifference). In the latter case, they often annotated the following utterance also with a combined label (e.g., {\fontfamily{cmtt}\selectfont hrt\_question}) as these tokens are more than just a hearer response token expressing continuous attention.  They labelled {\fontfamily{cmtt}\selectfont change\_hrt} when a short hearer response token expressed some attitude, when it occurred before a lapse in the conversation (i.e., when it was unclear who had the turn afterwards due to the very long following pause), in simultaneous speech and in one case because it was unclear if the previous turn by the other speaker ended in a functional question or not (“falls du das kennst” - 'if you know it', not embedded in a larger question; labelled as {\fontfamily{cmtt}\selectfont change\_question@}). They labelled {\fontfamily{cmtt}\selectfont hold\_hrt} in simultaneous speech and when a speaker produced a hearer response token before taking up the turn. They labelled {\fontfamily{cmtt}\selectfont question\_trail-off} when a question was asked but its utterance was not finished. This occurred, for instance, when a speaker did not finish formulating their question because the other speaker had projected the content of the question and had already started answering it.


\subsubsection{IPU annotation process}
\label{sec:IPUprocess}

Annotation of IPU was done in five stages (training, annotation by one annotator, correction by another annotator, evaluation and update of annotation criteria, final correction round). The first phase was the training phase for the segmentation and labelling of IPUs, in which the first author and three annotators held weekly meetings for several weeks. The annotators were students in General Linguistics.

After training, annotators did a first round of annotations. To avoid annotator-bias in the annotations, we divided five-minute extracts of each recording between the three annotators, and each annotator segmented and labelled 100 seconds. Then, one of the other annotators corrected these annotations. It took annotators approx. 60-90 minutes to make a first annotation and approx. 30-45 minutes to correct another annotator’s IPU annotations in 100 seconds of conversation.
We introduced a label to track uncertainties (\uncertain) and asked annotators to note down any insecurities and mark the unclear cases in the textgrid files. Some uncertainties were resolved in the correction by another annotator, some in the regular meetings with the first author, but 6.7\% of annotations with an uncertainty label remain at this stage. To have the uncertain cases explicitly tagged allows to either exclude them from an analysis, for instance when training a classifier to reduce noise in the data set, or to analyze them specifically.
Even after the training phase, we had regular meetings to discuss unclear cases, with decreasing frequency as annotators became more practiced. In these meetings, we solved many issues on a case-by-case basis with a consensus annotation. We collected and documented solved as well as unsolved unclear cases. That is, we collected unclear sequences with a note on what exactly the issue was and whether it could be solved or not. After several months of annotating, we sorted these issues according to topic and discussed them in the group to update the annotation criteria. For instance, we defined that repetitions of an interlocutor’s phrase or other longer, complex backchannels (e.g., "ah, im Kino, achso" - 'ah, at the cinema, I see') should be labeled as hearer response tokens despite their complexity, and we discussed the demarcation of questions vs. other kinds of turn-changes and decided that utterances explicitly eliciting information such as “du weißt das sicher” (‘you know this, don’t you’; when the current speaker does not have this information) should be labelled as \qu, even when they are not marked as questions by syntax or prosody\footnote{The English translation includes a tag question, but the German original does not. Instead, the discourse particle "sicher" ('surely') implies a more explicit listener orientation.}. This systematic discussion resulted in a revision of the annotation guidelines, which, in turn, resulted in an increased inter-rater reliability: a Cohen’s $\kappa$ of 0.59 before revision\footnote{One of the annotators had left the project before this stage. Therefore, we do not have any annotations by him to compare to after the revision. His Cohen’s $\kappa$ with the other two annotators before the revision was 0.54 and 0.56, respectively.}  vs. 0.84 after revision (cf., Section \ref{sec:IPUvalidation} for a more detailed description of inter-rater reliability and the most frequent confusions in IPU annotation). This change in annotation quality illustrates the importance of treating annotation as an iterative process. Accordingly, annotators performed another correction of all IPU annotations made before this evaluation, based on the new annotation criteria to ensure consistency.

\subsubsection{IPU validation}
\label{sec:IPUvalidation}

To validate the IPU annotations and to account for common confusions, we present a detailed evaluation of inter-annotator agreement of IPU annotations in this section.

\begin{figure}[ht]
    \centering
    \includegraphics[width=\linewidth]{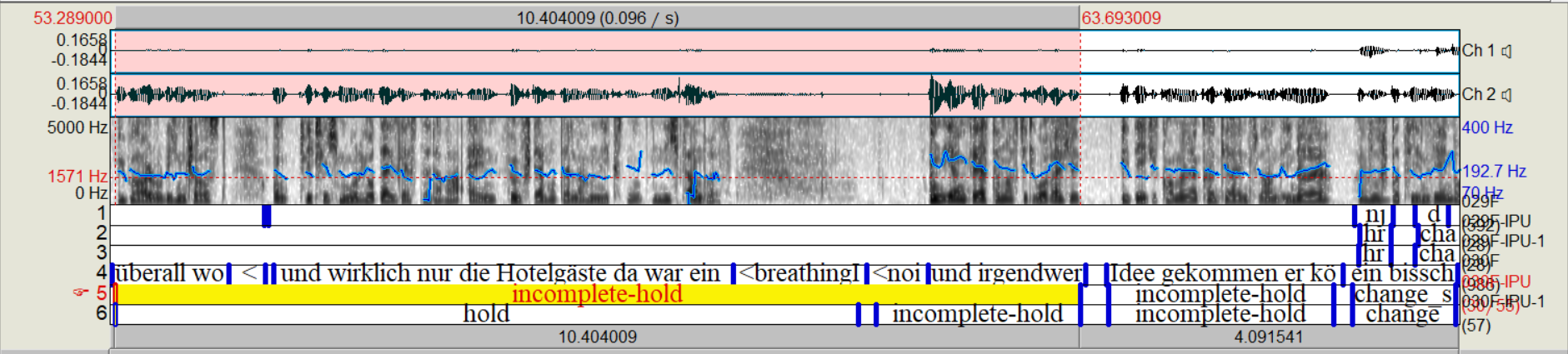}
    \caption{Praat screenshot of IPU annotations by two annotators (annotator 1: lines 2 and 5; annotator 2: lines 3 and 6), illustrating divergent IPU segmentation and its consequences for labelling. We combined the two annotators' annotations into the same Textgrid only for illustrative purposes, but they did not see each other's annotations in the annotation process.}
    \label{fig:Praat_IAA}
\end{figure}

We calculated Inter-Annotator Agreement of IPU labels on 300 seconds of conversation (100 sec in 3 conversations) annotated by two annotators. In these 300 sec of conversation, 135 IPUs were annotated, but in 11 cases (8\%), annotators did not agree on the presence of a pause, that is, one annotator set a boundary while the other did not. 
This happened when one annotator heard a breathing or other non-speech articulation (e.g., a nasal glottal stop) and did not interrupt the IPU while the other annotator did not hear the breathing, or heard a longer gap between breathing and speech, and annotated two separate IPUs. See, for instance, the selected interval in Figure \ref{fig:Praat_IAA}. One annotator segmented one IPU interval (line 5) while the other annotator set an intermediary boundary (line 6) and labelled the first interval as \hold{} and assigned the same label to the second interval (\ih) as the first annotator had to the whole interval.

Based on the 124 IPUs in which annotators set the same boundaries, Fleiss’s $\kappa$ is 0.84 (z = 19.7, p $>$ .0001), indicating near-perfect agreement \citep{LandisKoch1977}. That is, in 108 out of 124 observations, annotators were in agreement.
What Fleiss' $\kappa$ cannot capture, however, is when annotators were in partial agreement. In 5 cases, one annotator used a combined label which contained the label that the other annotator used (e.g., {\fontfamily{cmtt}\selectfont change\_hrt} vs. \change), and thus only in 11 cases, annotators were not in agreement at all. The confusions \change \hspace{0.5mm} vs. \hold \hspace{0.5mm} and \change \hspace{0.5mm} vs. \hrt \hspace{0.5mm} occurred in 3 cases each. In one \hold \hspace{0.5mm} vs. \change \hspace{0.5mm} case, the confusion resulted from a lapse, i.e., a very long pause, in the conversation after the labelled IPU \citep[cf.,][]{Sacks1974}. After this lapse, both speakers produced “hm” tokens, after which one speaker opened up a new topic, which made it unclear to the annotators whether the same speaker kept the turn after the lapse or whether the other speaker took it up. In another case, the confusion between \hold \hspace{0.5mm} and \change \hspace{0.5mm} resulted from a different analysis of the whole sequence, as in Example \ref{fp:meine004024}.
\vspace{-35pt}
\begin{fp}{fp:meine004024}{} 
\begin{center}
\includegraphics[width=\linewidth]{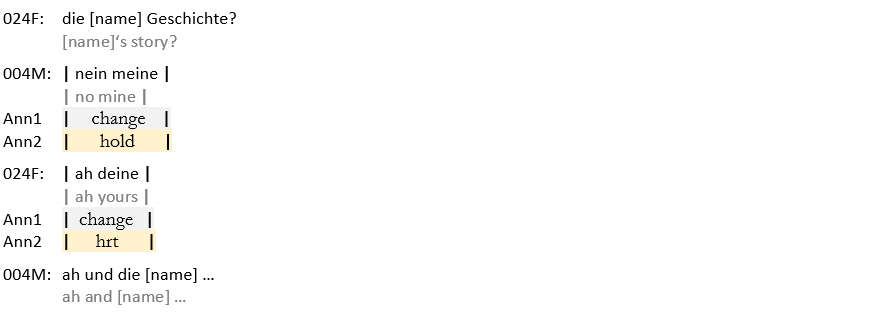}
\label{Example_meine004024}
\end{center}
\end{fp}
\vspace{-1.7em}

In this sequence, one annotator analysed 024F’s acknowledgement in "ah deine" as a turn in its own right, and, as a consequence, labelled the previous "nein meine" by 004M as \change. The other annotator analysed "ah deine" as a \hrt \hspace{0.5mm} and, as a consequence, labelled "nein, meine" as \hold, i.e., as part of the same turn as "ah und die [name]…" after 024F’s acknowledgement.

One confusion between \change \hspace{0.5mm} and \hold \hspace{0.5mm} was also due to a different analysis of the sequence, cf. Example \ref{fp:irgendwas001002}. 001M closed the utterance “irgendwas gibt’s da“ with a tag question. The two annotators interpreted this in different ways; one as an actual question that is answered by a short turn (“mhm”) labelled as \change, the other one as an utterance eliciting a backchannel, which was labelled as \hrt. In the second interpretation, the \hrt \hspace{0.5mm} does not interrupt the turn in progress by 001M. Thus, “irgendwas gibt’s da, ge?” was labelled as \hold.

\vspace{-1em}
\begin{fp}{fp:irgendwas001002}{} 
\begin{center}
\includegraphics[width=\linewidth]{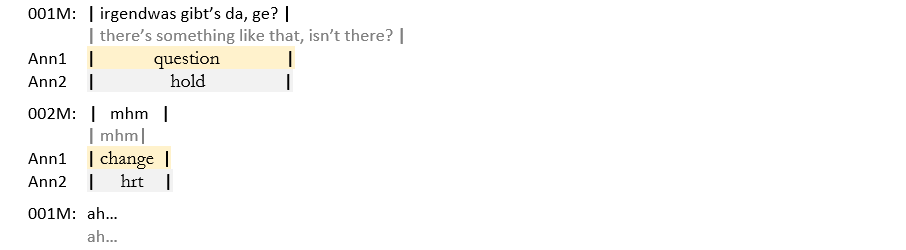}
\label{Example_irgendwas001002}
\end{center}
\end{fp}
\vspace{-1.7em}

When grouping the labels together into macro-categories (\textsc{turn-hold}: \hold, \ih; \textsc{turn-change}: \change, \qu, \selfint, \trailoff; \textsc{hrt}: \hrt), inter-rater agreement is equally high as for the individual labels ($\kappa$ = 0.84, z = 13.8, p $>$ .0001, N = 124). When grouping labels into \textsc{complete} (\hold, \change, \qu), \textsc{incomplete} (\ih, \selfint, \trailoff) and \textsc{hrt} (\hrt), the inter-rater agreement rises ($\kappa$ = 0.875, z = 13.5, p $>$ .0001, N = 124). 

In conclusion, the inter-rater agreement of our IPU turn-taking annotations is within the range of studies that applied \citeauthor{gravanohirschberg2011}'s \citeyearpar{gravanohirschberg2011} annotation system to IPUs in languages other than English (Spanish: 0.88; and Slovak: 0.81 \citep{Brusco2020}; compared to a higher agreement in English: 0.99 \citep{gravanohirschberg2011, Levitan2015}, and 0.91 \citep{Brusco2020}).

\subsection{Points of potential completion (PCOMP)}
\label{sec:PCOMP}

\subsubsection{Domain of PCOMP annotations}
\label{sec:PCOMPdomain}
The second layer on which turn-taking was annotated was in the domain of potential completion (PCOMP). For segmentation on this layer, we followed \citeauthor{Clancy1996}'s \citeyearpar{Clancy1996} definition of ‘grammatical completion’:

\setlength{\leftskip}{1cm}
\noindent
``We judged an utterance to be grammatically complete if, in its sequential context, it could be interpreted as a complete clause, i.e., with an overt or directly recoverable predicate, without considering intonation. In the category of grammatically complete utterances, we also included elliptical clauses and answers to questions. […] A grammatical completion point, then, is a point at which the speaker could have stopped and have produced a grammatically complete utterance, though not necessarily one that is accompanied by intonational or interactional completion" \cite[pp. 336f.]{Clancy1996}.

\setlength{\leftskip}{0pt}
\noindent
In previous studies, such points in speech have also been called ‘points of possible syntactic completion’ \citep[SYNCOMP;][]{LocalWalker2012} and ‘potential turn boundaries’ \citep[PTB;][]{Zellers2017}. This notion overlaps to a large degree with ‘transition relevance places’ \citep[TRP;][]{Sacks1974}. In most of the Conversation Analysis literature, however, TRPs are generally not only characterized by grammatical completeness, but also by prosodic completeness. In line with \cite{Clancy1996}, \cite{LocalWalker2012} and \cite{Zellers2017}, we did not include prosodic completion into the definition of these points for our annotation system. While native speakers may have an intuition about what is prosodically complete, different listeners might interpret prosody in different manners. Which phonetic parameters indicate prosodic completion in Austrian German is still a topic of investigation (e.g., a phrase-final rising intonation could indicate several different things) and there are no clear criteria yet for determining completion in this variety. Therefore, we followed these authors’ purely syntactic-pragmatic approach to segmentation.
To be considered potentially complete, an utterance needs a predicate (including its complements), either expressed in the utterance itself or recoverable from the previous context. In the invented example in \ref{fp:Buch} a) below, a boundary (indicated by a vertical line) is set after Buch (‘book’). Even though the speaker continues speaking, “sie schenkt ihm ein Buch” could be a complete sentence because it includes the verb (“schenken”) and its obligatory direct and indirect objects (“Buch” and “ihm”). Additional boundaries are added after each increment (“zum Geburtstag”, “morgen”).
German syntax allows the insertion of several constituents between a transitive verb and its direct object. In Example \ref{fp:Buch} b), for instance, the increments from a) are inserted before the direct object. Therefore, no boundary is set until the direct object is expressed.

\begin{fp}{fp:Buch}{} 
\begin{center}
\includegraphics[width=\linewidth]{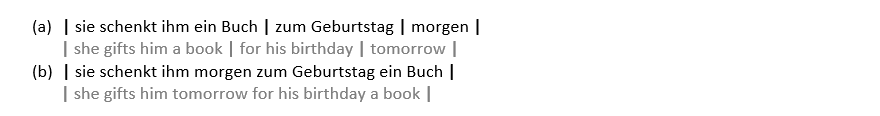}
\label{Example_Buch}
\end{center}
\end{fp}

The same boundary criteria apply to separable predicates, such as, present perfect with participle II (e.g., “ich habe …. gegessen” – ‘I ate …’/’I have eaten…’) and separable verb prefixes (e.g., “ich ziehe … an” – ‘I put … on’). No boundary is set in these cases until all required elements (e.g., the participle and the separated prefix) have been expressed. If a main clause follows a subordinate clause, a boundary is set only when the main clause has been expressed. However, turn projection strategies with a scope wider than the sentence, such as “first, … second, …”, are not considered for PCOMP segmentation. That is, even if more talk on a second issue is projected by “first, …”, the maximum domain of potential completion is still the sentence.
Many utterances in a conversation, however, are not full sentences containing a predicate. Therefore, the preceding context is also considered, in particular previous turns by another speaker. In the invented Example \ref{fp:Stadtpark}, several answer options are given to a question asking for a location. Potential completion is reached when a location is expressed in the answer, even if the answer does not contain a predicate, because it is recoverable from the question in the previous turn. Thus, a boundary is set as soon as the location is given, as in (a) and (b). Even if an answer is given in a longer format, for instance, by a main clause including additional information, as in (c), no boundary is set before the location is expressed. The same boundary is set when the predicate is expressed again after the location, as in (d), since the location alone could be a complete answer.

\vspace{-1em}
\begin{fp}{fp:Stadtpark}{} 
\begin{center}
\includegraphics[width=\linewidth]{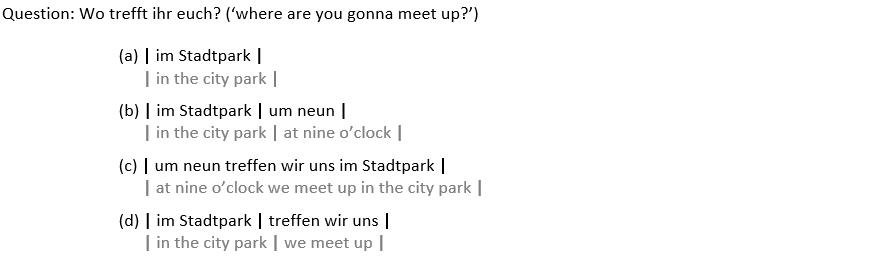}
\label{Example_Stadtpark}
\end{center}
\end{fp}
\vspace{-1.7em}

In contrast to segmentation on the IPU layer, breathings and other non-speech articulations are not included into annotated PCOMP intervals (cf., Figure \ref{fig:Praat}). After a pause (e.g., turn-initially), a segmented interval starts when speech starts and ends when the first PCOMP is reached. No boundaries are set at pauses before a PCOMP is reached. Example \ref{fp:013014auskennst} illustrates how segmentation differs between the PCOMP and IPU layer when a speaker pauses at a point of maximum grammatical control \citep{schegloff1996turn}. On the IPU layer, the pause is segmented regardless of grammatical projection. On the PCOMP layer, a boundary is set before “aber” (‘but’), and the pause is inside the next PCOMP interval, which only ends when the sentence is complete.
\vspace{-1em}
\begin{fp}{fp:013014auskennst}{} 
\begin{center}
\includegraphics[width=\linewidth]{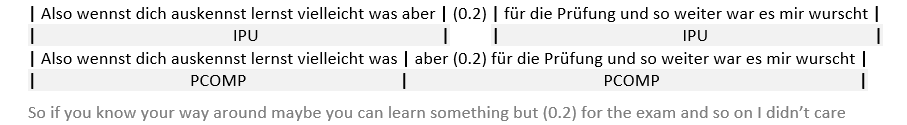}
\label{Example_013014auskennst}
\end{center}
\end{fp}

While PCOMP refers to end points rather than units, in the remainder of this paper, we still refer to the PCOMP annotations as intervals that start when speech starts at the beginning of a turn (or hearer response token), or at the end boundary of an immediately preceding PCOMP if more speech is produced, and end when a PCOMP is reached (cf., grey intervals in Examples \ref{fp:013014auskennst}-\ref{fp:029030Vaeter}).

In contrast to the IPU layer, where setting boundaries and assigning labels can be done independently from each other (cf., Section \ref{sec:IPUdomain}), on the PCOMP layer, boundaries are intricately linked to the labels, since they are based on syntactic-pragmatic criteria. Thus, a change in boundary criteria (e.g., based on prosodic instead of syntactic domains) would result in a totally different phenomenon being annotated.

\subsubsection{PCOMP labels}
\label{sec:PCOMPlabels}

\begin{figure}[ht]

  \begin{adjustbox}{addcode={\begin{minipage}{\width}}{\caption{%
      Decision tree for assigning single PCOMP labels and the \coll label. A decision tree for assigning PCOMP single as well as frequent combined labels is presented in Figure \ref{fig:Flowchart_PCOMPcombined} in \nameref{sec:appendixC}. For comparability, branches in this figure are coloured in the same colours as the PCOMP intervals in Figure \ref{fig:dynamics028008_PCOMP} and the branches in Figure \ref{fig:Flowchart_PCOMPcombined}. }\label{fig:Flowchart_PCOMPsimple}\end{minipage}},rotate=90,center}     
      \includegraphics[scale=.7]{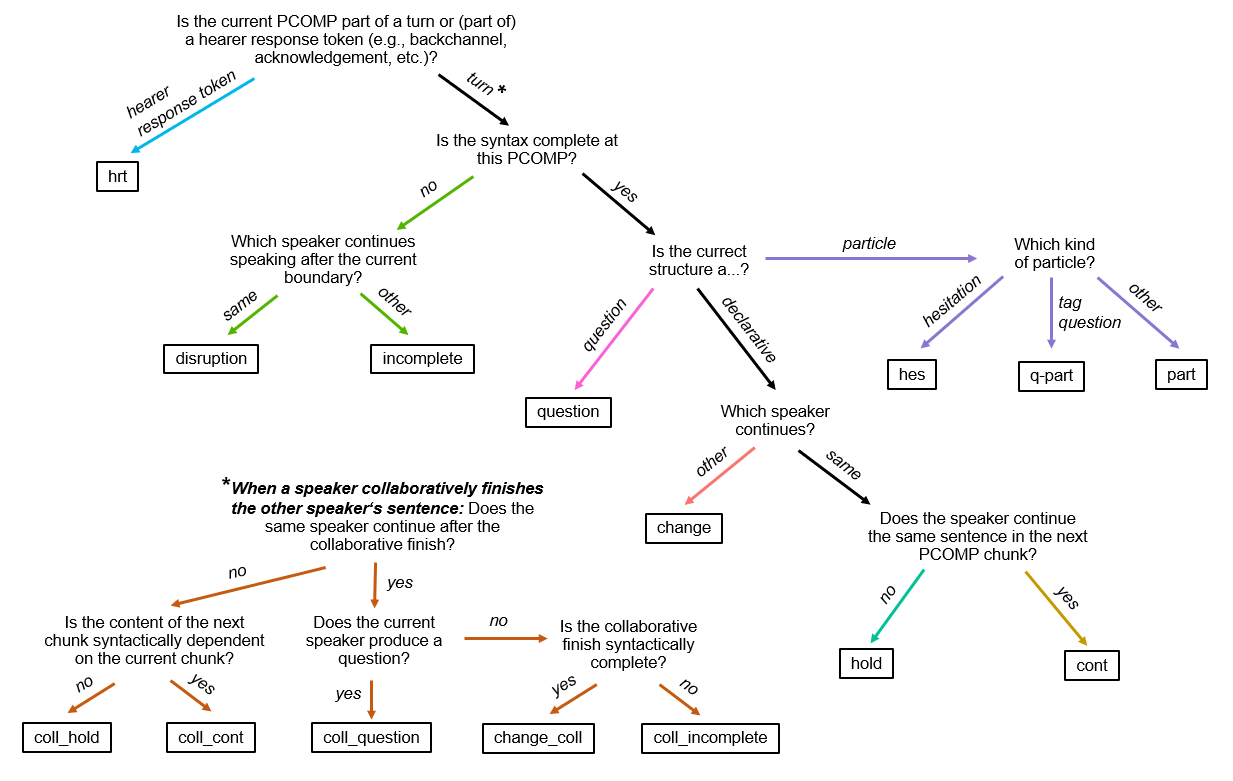}%
  \end{adjustbox}
\end{figure}

PCOMP intervals received one of eleven labels. An overview of all possible labels assigned on the PCOMP layer in GRASS are presented in Table \ref{tab:PCOMPlabels}, and Figure \ref{fig:Flowchart_PCOMPsimple} presents a decision tree for assigning single PCOMP labels\footnote{A decision tree for assigning single as well as combined PCOMP labels is presented in Figure \ref{fig:Flowchart_PCOMPcombined} in \nameref{sec:appendixC}.}. PCOMP labels combine syntactic and turn-taking criteria:

\begin{itemize}
  \item \hold, when the current speaker continues after the PCOMP and starts a new syntactic structure:

\vspace{-1.7em}
\setlength{\leftskip}{1cm}
  \begin{fp}{fp:Kino1}{} 
  \begin{center}
    \includegraphics[width=\linewidth]{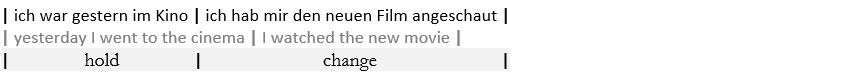}
    \label{Example_Kino1}
  \end{center}
  \end{fp}
\setlength{\leftskip}{0cm}
\vspace{-2.5em}

  \item \cont, when the current speaker continues after the PCOMP and continues the syntactic structure with increments:

\vspace{-1.7em}
\setlength{\leftskip}{1cm}
  \begin{fp}{fp:Kino2}{} 
  \begin{center}
    \includegraphics[width=\linewidth]{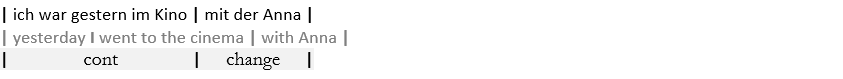}
    \label{Example_Kino2}
  \end{center}
  \end{fp}
\setlength{\leftskip}{0cm}
\vspace{-2.5em}

  \item \change, when the current PCOMP is followed by a turn change, that is, the current interval is the last one in the turn and the interlocutor takes up the turn next:

\vspace{-1.7em}
\setlength{\leftskip}{1cm}
  \begin{fp}{fp:Kino3}{} 
  \begin{center}
    \includegraphics[width=\linewidth]{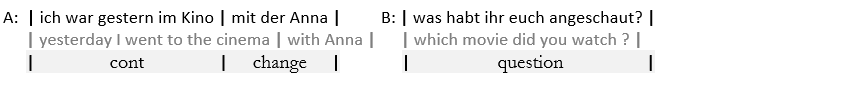}
    \label{Example_Kino3}
  \end{center}
  \end{fp}
\setlength{\leftskip}{0cm} 
\vspace{-2.5em}

  \item \particle, for syntactically independent discourse particles, such as “ja”, “halt”\footnote{“halt” can be used, for instance, to mark information as uncontroversial/topical \citep{Trotzke2020}, or to mark exasperation \citep{Zimmermann2012}.}, “weißt” (‘you know’), that are uttered after a PCOMP boundary or at the beginning of a turn. This label is not used when, for instance, “ja” is used as an answer to a question, but it is applied when “ja” is used as an upbeat to a turn (e.g., “ja es ist weil du nie Dexter schaust“ in Example \ref{fp:PCOMP_dynamics} text example). Discourse particles are not segmented unless they are uttered turn-initially or at a PCOMP. For instance:

\vspace{-1.7em}
\setlength{\leftskip}{1cm}
  \begin{fp}{fp:Kugelhalt}{}
  \begin{center}
    \includegraphics[width=\linewidth]{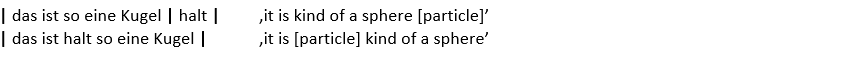}
    \label{Example_Kugelhalt}
  \end{center}
  \end{fp}
\setlength{\leftskip}{0cm}
\vspace{-2.5em}

  \item \qupart, for question particles (tag questions) that can transform a declarative to a question or elicit backchannels, such as “ne”, “nicht”, “oder”, “ge/gell” (comparable to English “right” and “isn’t it”).
  \item \qu: in contrast to the \qu \hspace{0.5mm} label on the IPU layer, where only questions that are responded to by the interlocutor are annotated as \qu, on the PCOMP layer any utterance that is syntactically and/or prosodically  
  marked as a question is labelled as \qu. The unmarked question is turn-yielding. If, however, the same speaker continues speaking after a question, for instance, in a series of questions, each PCOMP interval is labelled as {\fontfamily{cmtt}\selectfont hold\_question} or {\fontfamily{cmtt}\selectfont cont\_question} (cf., Section \ref{sec:PCOMPcombined} and Example \ref{fp:wohnen001002}).
  \item \hes, for hesitation particles like “eh” and “ehm” that are uttered after a PCOMP or at the beginning of a turn. Similar to discourse particles, hesitation particles are not segmented if they occur in the middle of a construction (e.g., ‘it was ehm I think yesterday’).
  \item \coll, when a speaker collaboratively finishes another speaker’s sentence. This label looks backwards (i.e., at how the current utterance is related to what came before) rather than forwards (i.e., at how the current utterance is related to what comes next). Therefore, collaborative finishes always require a combined label, for instance, {\fontfamily{cmtt}\selectfont coll\_hold} if the speaker producing the collaborative finish continues talking, or {\fontfamily{cmtt}\selectfont coll\_change} if the other speaker continues speaking after the collaborative finish.
  \item \hrt: same as \hrt \hspace{0.5mm} on the IPU layer, except for complex hearer response tokens that contain several PCOMPs. In these cases, boundaries were set according to PCOMP criteria, and the respective interval received a double label (e.g., {\fontfamily{cmtt}\selectfont hold\_hrt}; cf., V. in Section \ref{sec:PCOMPcombined}).
\end{itemize}

To capture two additional phenomena where speakers produce syntactically-pragmatically incomplete structures, we introduced two cases in which a boundary should be set even though a speaker had not reached a point of potential completion:

\begin{itemize}
  \item \disruption: the current speaker disrupts a syntactic structure in progress to start a new syntactic structure, that is, to rephrase themselves (Example \ref{fp:zuFuss}). Examples \ref{fp:Keplerbruecke} and \ref{fp:Platine} illustrate how this annotation allows us to disambiguate between \ih s (cf., IPU label description in Section \ref{sec:IPUlabels}) after which the same structure is continued vs. ones where a new sentence is started after the pause. In the former case, the utterance before the pause is not segmented into a separate PCOMP (Example \ref{fp:Keplerbruecke}), while in the latter case, it is labelled as \disruption{} on the PCOMP layer (Example \ref{fp:Platine}). 
  
  The \disruption{} label also applies to self-interruptions with subsequent rephrasing without a pause. We did not, however, annotate \disruption \hspace{0.5mm} when speakers just repeated part of a phrase and then continued their syntactic construction (e.g., “der war der war gestern da” – ‘he was he was here yesterday”).

\vspace{-1.7em}
\setlength{\leftskip}{1cm}
  \begin{fp}{fp:zuFuss}{} 
  \begin{center}
    \includegraphics[width=\linewidth]{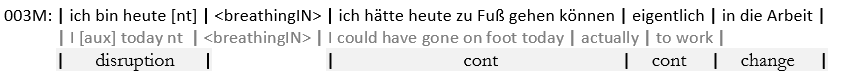}
    \label{Example_zuFuss}
  \end{center}
  \end{fp}
\vspace{-2em}

\vspace{-1.7em}
  \begin{fp}{fp:Keplerbruecke}{} 
  \begin{center}
    \includegraphics[width=\linewidth]{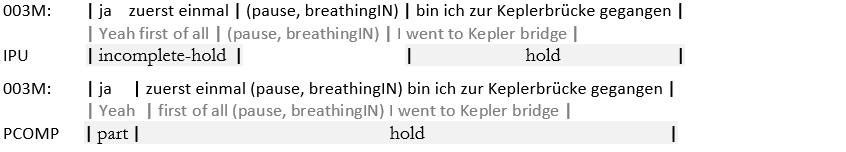}
    \label{Example_Keplerbruecke}
  \end{center}
  \end{fp}
\vspace{-2em}

\vspace{-1.7em}
  \begin{fp}{fp:Platine}{} 
  \begin{center}
    \includegraphics[width=\linewidth]{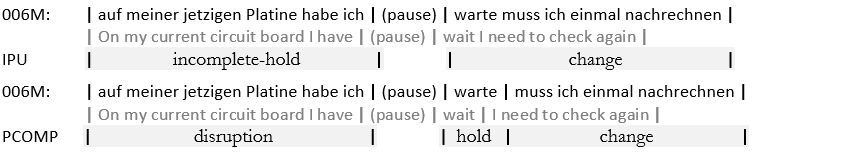}
    \label{Example_Platine}
  \end{center}
  \end{fp}
  \setlength{\leftskip}{0cm}
\vspace{-2em}

  \item \incomplete: the current speaker produces an incomplete structure before a turn change. That is, the other speaker takes up the turn even though the current speaker has not reached a point of potential completion (e.g., trail-offs (Walker 2012) or self-interruptions to cede the turn to the interlocutor).
\end{itemize}

Table \ref{tab:PCOMPcounts} presents the distribution of single PCOMP labels in different conversations in GRASS.
Overall, 3771 PCOMP intervals were annotated in 70 minutes of conversation in GRASS (5 minutes in 14 conversations; a subset of the data annotated on the IPU layer). 3415 PCOMPs (91\%) were labelled with single labels, 316 PCOMPs (8\%) with combined labels (cf., Section \ref{sec:PCOMPcombined}), and 47 PCOMPs (1.2\%) were labelled as uncertain.

\clearpage

\begin{table}
\begin{center}
\caption{List of PCOMP labels and their definitions.}
\begin{tabular}{p{0.2\textwidth}p{0.8\textwidth}}
  \hline
 \textbf{PCOMP label} & \textbf{Definition} \\  \hline
 \hold &  same speaker continues speaking after the PCOMP by starting a new sentence\\  \hline
 \cont &  same speaker continues speaking after the PCOMP by continuing the same sentence with the addition of increments\\  \hline
 \change & other speaker continues speaking after the current speaker reaches a PCOMP \\   \hline
 \particle & discourse particle uttered after a PCOMP or at the beginning of a turn \\   \hline
 \qupart & question particle (tag question) that transforms a declarative utterance into a question or is used to elicit some kind of listener feedback (e.g., a backchannel) \\   \hline
 \qu & syntactic and/or prosodic question \\   \hline
 \hes & hesitation particle uttered after a PCOMP or at the beginning of a turn \\   \hline
 \coll & current speaker collaboratively finishes the previous speaker’s sentence. This label is always combined with another label to indicate whether the same or the other speaker continues speaking after the PCOMP (cf., Section \ref{sec:PCOMPcombined} \\   \hline
 \hrt & hearer response token; usually short backchannels, continuers, acknowledgements, etc., that do not contain a (new) proposition of their own and do not take up the turn \\   \hline
 \disruption & current speaker does not reach a PCOMP, but interrupts themselves to rephrase and start a new sentence \\   \hline
 \incomplete & current speaker does not reach a PCOMP before the other speaker takes up the turn \\   \hline
 \combined & combination of two of the labels above (in alphabetical order). Common cases are described in Section \ref{sec:PCOMPcombined} \\   \hline
 \uncertain & indicates uncertainty about a label (cf., Section \ref{sec:PCOMPprocess}); may also co-occur with a combined label to indicate the uncertainty between two specific labels (cf., Section \ref{sec:PCOMPvalidation}) \\       \hline
\end{tabular}
\label{tab:PCOMPlabels}
\end{center}
\end{table}

A much higher number of turn-holding annotations (\hold: N = 1147; \cont: N = 419) than turn-yielding annotations (\change: N = 338; \qu: N = 119; \incomplete: N = 124) tells us that most turns consist of more than one PCOMP interval. 
Assuming that the annotations \change, \qu, \incomplete{} and \hold{} indicate the end of turn construction units (TCU)\footnote{\cont{} indicates that the speaker continues the \textit{same} TCU in the next interval, and particles, tag questions and hesitations are not TCUs in their own right.}, a ratio of 1,147 turn-holding PCOMPs (\hold) to 581 turn-yielding PCOMPs (\change, \qu{} and \incomplete) indicates that turns in our data consist, on average, of about three TCUs\footnote{We are aware that this is an oversimplification of the concept of TCUs \citep[cf., ][]{Selting2000}, since we neither considered prosody nor multi-unit turn projection for our concept of PCOMP.}. The distribution of PCOMP labels in Table \ref{tab:PCOMPcounts} illustrates that interlocutor’s turn-internal structures are more similar in some conversations (e.g. 001M002M) than in others. For instance, 015M has a ratio of turn-holding to turn-yielding PCOMPS of 137:52 while 017M has a ratio of 32:38. This indicates that 015M tends to produce longer turns that, on average, contain three to four PCOMPs (e.g., $\vert$ \hold{} $\vert$ \cont{} $\vert$ \hold{} $\vert$ \change{} $\vert$), while 017M tends to produce shorter turns that contain only one or two PCOMPs (e.g., $\vert$ \hold{} $\vert$ \change{} $\vert$ or only $\vert$ \change{} $\vert$). Moreover, 015M produced about four times as many {\disruption}s and many \hes, indicating that his speech is more disfluent than his interlocutor’s.

\begin{table}[ht]
\centering
\caption{Distribution of single PCOMP annotations per speaker, sorted by conversation. The label \coll{} is not listed here, since it always requires a combined label (cf., Section \ref{sec:PCOMPlabels} and Table \ref{tab:PCOMP_combined}).}
\begin{tabular}{lrrrrrrrrrr}
  \hline
spkID & hold & cont & change & part & q-part & question & hes & hrt & disruption & incomplete \\ 
  \hline
001M &  39 &   5 &  20 &  14 &   4 &   3 &   1 &  39 &   7 &   5 \\ 
  002M &  37 &   4 &  26 &  16 &   1 &   6 &   2 &  44 &   6 &   1 \\ \hline
  003M &  37 &  17 &  17 &   3 &   1 &   2 &   3 &   7 &  10 &  \\ 
  023F &  30 &   8 &   9 &   4 &   1 &  12 &  &   5 &   2 &  \\  \hline
  006M &  31 &   9 &   8 &   8 &   4 &   5 &   1 &  18 &  10 &   3 \\ 
  007M &  39 &  11 &  12 &   7 &   4 &   2 &  &  10 &   3 &   2 \\  \hline
  009M &  41 &   8 &  12 &  15 &   3 &   5 &  &  23 &   9 &   5 \\ 
  010M &  66 &  14 &  12 &   8 &   9 &   1 &   3 &  14 &  18 &   4 \\  \hline
  013M &  24 &  12 &   7 &   9 &   4 &   7 &  &  42 &  &   3 \\ 
  014M &  59 &  23 &  15 &  11 &   1 &  &   1 &  22 &   2 &   5 \\  \hline
  015M & 103 &  34 &  33 &  15 &  &   1 &  13 &  15 &  20 &  18 \\ 
  017M &  25 &   7 &   6 &  15 &  20 &  20 &  &  32 &   4 &  12 \\  \hline
  021F &  29 &  14 &  13 &   7 &  &   2 &  &  42 &   5 &   5 \\ 
  022F &  40 &  18 &   7 &   9 &   1 &   3 &  &  24 &  17 &   7 \\  \hline
  005M &  26 &  15 &   6 &  13 &   6 &   6 &   1 &  31 &   6 &   5 \\ 
  025F &  38 &  19 &  10 &   6 &   1 &   4 &   1 &  15 &   2 &   2 \\  \hline
  026F & 103 &  34 &   6 &  32 &   5 &   2 &   1 &  18 &  19 &  10 \\ 
  027F &  22 &   7 &   7 &   3 &   8 &   1 &  &  62 &   4 &   5 \\  \hline
  008M &  31 &   8 &  16 &  11 &   1 &   9 &   2 &   4 &   4 &  \\ 
  028F &  39 &   7 &  15 &  12 &  &   2 &   1 &  11 &   4 &  10 \\  \hline
  029F &  14 &  10 &   9 &   4 &  &   1 &   1 &  11 &  &   1 \\ 
  030F &  53 &  24 &   9 &   5 &   1 &   2 &  &   7 &   9 &   1 \\  \hline
  011M &  57 &  44 &  12 &  15 &  &   4 &  &   9 &   9 &   2 \\ 
  031F &  22 &  14 &  10 &   7 &  &   7 &   1 &  41 &   6 &   7 \\  \hline
  012M &  35 &  19 &  20 &  17 &   4 &  &  &  21 &   7 &   2 \\ 
  032F &  33 &  17 &  11 &  11 &   3 &   6 &  &   6 &   4 &   3 \\  \hline
  038F &  57 &  14 &   8 &  10 &  &   1 &   8 &   6 &  23 &   4 \\ 
  039F &  17 &   3 &   2 &   6 &   2 &   5 &  &  52 &   3 &   2 \\  \hline
  TOTAL & 1147 & 419 & 338 & 293 &  84 & 119 &  40 & 631 & 213 & 124 \\ 
   \hline
\end{tabular}
\label{tab:PCOMPcounts}
\end{table}

\subsubsection{PCOMP combined labels}
\label{sec:PCOMPcombined}

To account for the full range of interlocutors’ behaviours, we allowed for the combination of labels also on the PCOMP layer when more than one label applied. Note that, analogous to IPU labels, combined PCOMP labels are different from uncertain annotations, which are indicated by \uncertain. Depending on the research question, combined PCOMP labels easily allow for their exclusion from the data set, for their grouping with either one or the other label, or for their investigation, specifically. 

For PCOMP, combined labels are more integral to the annotation process than for IPUs. Since PCOMP labels are defined by a combination of turn-taking and syntactic criteria, some syntactically defined labels had a default turn-taking setting derived from their prevalence in a particular position. For instance, questions are most often turn-yielding and hesitations and other discourse particles (except for \qupart) are most often turn-holding. However, the latter can also occur in turn-final, as questions can occur in turn-medial position. In these cases, they received a combined label ({\fontfamily{cmtt}\selectfont change\_hes}, {\fontfamily{cmtt}\selectfont change\_part}, {\fontfamily{cmtt}\selectfont hold\_question}). Moreover, some combined labels result from complex behaviour, such as, overlapping speech after a TRP which the interlocutor interprets as turn-yielding, while the current speaker continues speaking (annotated as {\fontfamily{cmtt}\selectfont change\_hold}). The \coll{} label is backwards-looking, so collaborative finishes always received a combined label that also indicated whether they were, for instance, turn-holding or turn-yielding (cf., Table \ref{tab:PCOMP_combined}). Below, we present a more detailed analysis of frequent combined labels. In Figure \ref{fig:Flowchart_PCOMPcombined} in \nameref{sec:appendixC}, we provide guidelines about when combined labels should be assigned in the form of a decision tree.

In comparison to the IPU annotation layer, combined labels have a higher frequency on this layer (N = 316; 8\% of PCOMP intervals vs. 1.8\% of IPU intervals). 
34 different combined labels were annotated. Table \ref{tab:PCOMP_combined} lists combined labels that were assigned more than four times, as well as combined \coll{} labels. 

\begin{table}[ht]
\centering
\caption{Combined PCOMP labels that occurred more than four times and combined \coll{} labels.}
\vspace{+10pt}
\begin{tabular}{lr}
  \hline 
Combined label & number of occurrence \\ 
  \hline  
  change\_hold &  48 \\ 
  change\_hrt &   5 \\  
  change\_part &  21 \\  
  change\_question &   9 \\ 
  cont\_hrt &   4 \\ 
  cont\_question &  50 \\ 
  disruption\_incomplete &   9 \\  
  disruption\_question &   4 \\ 
  hold\_hrt &  15 \\ 
  hold\_question &  64 \\  
  hrt\_part &  19 \\  
  hrt\_question &  13 \\  
  incomplete\_question &  14 \\ 
   \hline 
     change\_coll &   4 \\ 
  coll\_cont &   1 \\  
  coll\_hold &   2 \\  
  coll\_incomplete &   1 \\ 
  coll\_question &   4 \\ 
  \hline
\end{tabular}
\label{tab:PCOMP_combined}
\end{table}
\newpage
Eight combined labels, which can be grouped into five conditions (I to V), occurred more than ten times each:

I) {\fontfamily{cmtt}\selectfont change\_hold} (n = 48) was labelled when both interlocutors started talking at the same time after the PCOMP, 
and when the interlocutor responded in overlap while the current speaker continued talking, 
cf., "ich wollte Lärche einbauen" in Example \ref{fp:Boot}\footnote{Cf., Example \ref{fp:Stadtpark} d) for the boundary after “ich wollte”, which also carries a contrastive accent here.}.

\vspace{-2em}
\begin{fp}{fp:Boot}{}
  \begin{center}
    \includegraphics[width=\linewidth]{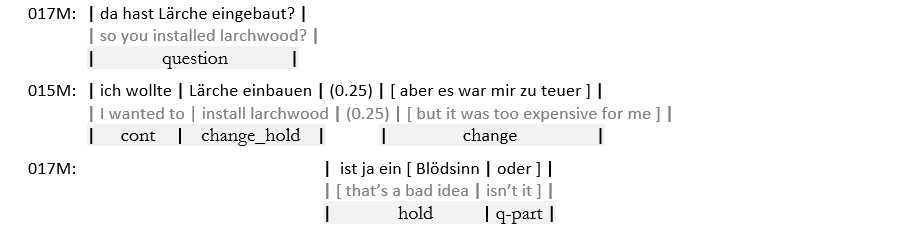}
    \label{Example_Boot015017}
  \end{center}
\end{fp}
\vspace{-2em}

II) Discourse particles that occurred at the end of a turn were labelled as {\fontfamily{cmtt}\selectfont change\_part} (n = 21; e.g., “ja”, “und so”, “oder so”, “halt”, “eh”; e.g., in “das ist so eine Kugel halt” – ‘it’s a kind of sphere [particle]’ if another speaker takes up the turn afterwards). 

III) When a speaker asked a question and the interlocutor answered before the formulation of the question was complete, the label {\fontfamily{cmtt}\selectfont incomplete\_question} was applied (n = 13). In most cases, the next speaker came in early with their answer, so the current speaker asking the question interrupted themselves to cede the turn (Example \ref{fp:zufrieden038039}\footnote{Judging from the context, the most likely word intended here is “frustrating”, but it was not realized because 039F interrupted herself to yield the floor to 038F, who had started answering 039F’s question at that point.}), or the question ended in “oder…”, often before a pause, which cannot be characterised as a question particle but as a conjunction. 

\vspace{-1em}
\begin{fp}{fp:zufrieden038039}{}
  \begin{center}
    \includegraphics[width=\linewidth]{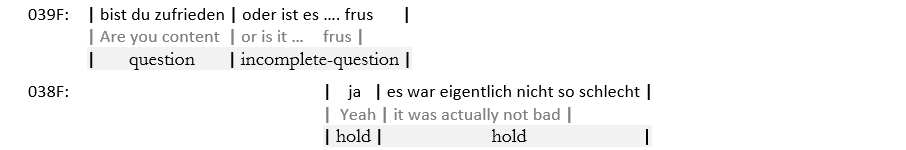}
    \label{Example_zufrieden038039}
  \end{center}
\end{fp}
\vspace{-2em}

IV) Speakers often produce several questions in sequence or added increments to their questions, resulting in several questioning PCOMP intervals that still hold the turn. Example \ref{fp:wohnen001002} illustrates this behaviour.

\vspace{-1em}
\begin{fp}{fp:wohnen001002}{}
  \begin{center}
    \includegraphics[width=\linewidth]{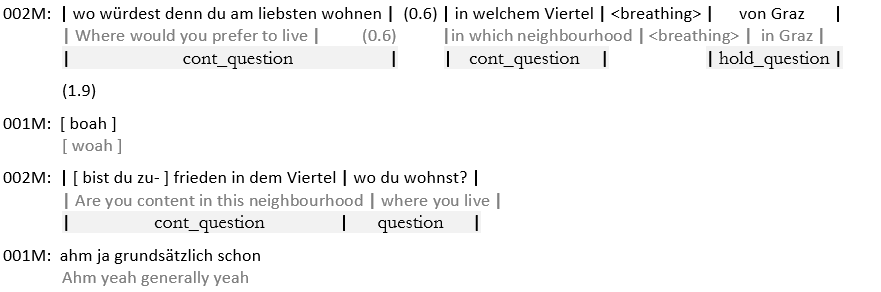}
    \label{Example_wohnen001002}
  \end{center}
\end{fp}
\vspace{-2em}

In Example \ref{fp:wohnen001002}, 002M asks a question in the first PCOMP interval that is not immediately answered. He thus adds increments to the initial question, and after a long pause of 1.9 sec he asks a related, but new question which is answered. Except for the last PCOMP, which is labelled as \qu \hspace{0.5mm} (because it transfers the turn to 001M), all previous parts of such sequences that try to transfer the turn are labelled as {\fontfamily{cmtt}\selectfont hold\_question} (n = 64) or {\fontfamily{cmtt}\selectfont cont\_question} (n = 50), depending on whether the speaker continues the same sentence after the PCOMP or not (cf., definition of \hold \hspace{0.5mm} vs. \cont).

V) Some of the longer hearer response tokens that contain more than one PCOMP require combined labels. For instance, a hearer response like “na $\vert$ auch nicht viel $\vert$ ja $\vert$ stimmt $\vert$” ( ‘no $\vert$ also not a lot $\vert$ yeah $\vert$ right’ ) contains four PCOMPs. “na” and “ja” were labelled as {\fontfamily{cmtt}\selectfont hrt\_part} (n = 19) and “auch nicht viel” was labelled as {\fontfamily{cmtt}\selectfont hold\_hrt} (n = 15). Moreover, some hearer response tokens have a decidedly questioning character. For instance, “echt” with a rising question intonation as well as repetitions of something the interlocutor just said with a rising question intonation (e.g., “nein?” - ‘no?’, “Wasser?” - ‘water?’) prompt the interlocutor to elaborate more on the same topic and were labelled as {\fontfamily{cmtt}\selectfont hrt\_question} (n = 13).

\subsubsection{PCOMP annotation process}
\label{sec:PCOMPprocess}

\begin{figure}[ht]
  \begin{adjustbox}{addcode={\begin{minipage}{\width}}{\caption{%
      Decision tree for assigning IPU labels, including single and frequent double labels. The label {\fontfamily{cmtt}\selectfont change\_hrt} is not included here, because it was assigned to various diverse behaviours that cannot easily be captured by a single question, such as described in Section \ref{sec:IPUcombined}.
      }\end{minipage}},rotate=90,center}
      \includegraphics[scale=.7]{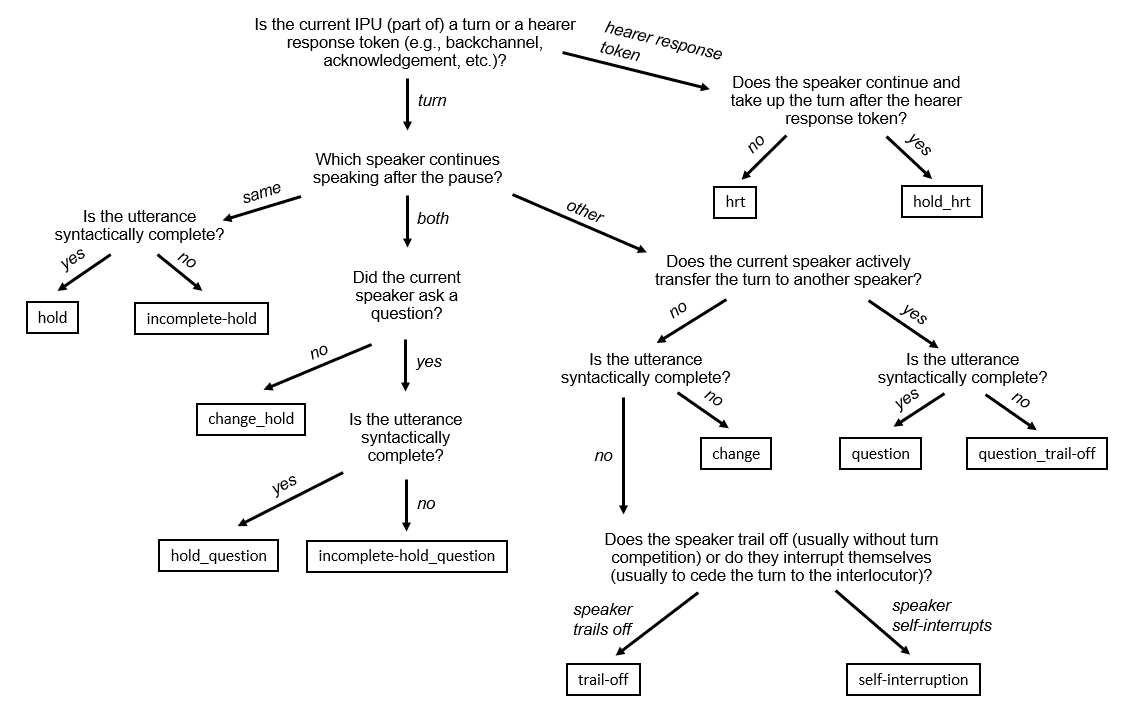}%
  \end{adjustbox}
  \label{fig:Flowchart_IPUcombined}
\end{figure}

PCOMP annotations were done after IPU annotations. Annotators saw the IPU annotations on a separate tier in Praat while annotating PCOMP, along with the orthographic transcription of the utterances.
The annotation of PCOMP requires a firmer background in syntax as well as experience with phonetic segmentation because boundaries had to be set with precision not only at pauses (as for IPUs), but also within the speech stream, that is, between two phones. The amount of reduction and coarticulation in conversational speech makes this task particularly challenging. The training of someone without any background in syntax or phonetic segmentation would have taken an excessive amount of time. However, one of the annotators was specialising in morpho-syntax in her studies at that time and had already had some experience with phonetic segmentation. Therefore, we decided that she would do the first round of annotations for all recordings, that is, the segmentation and a first round of labelling, while the other annotator did the corrections of all recordings. The greater complexity and difficulty of PCOMP annotation in comparison to IPU annotation is also reflected in the more complex decision tree for annotating single and combined labels in Figure \ref{fig:Flowchart_PCOMPcombined} (PCOMP annotation) in comparison to Figure \ref{fig:Flowchart_IPUcombined} (IPU annotation).

In most recordings, it took the first annotator approx. 60-90 minutes to make a first annotation of 100 seconds (including self-correction after some time), though in some cases, the first round took up to three hours for 100 seconds of conversation. This occurred when speakers tended to use complex syntactic structures or a lot of elliptic structures, if they were unintelligible in some parts due to, e.g., strong reduction and unclear articulations, sometimes in combination with dialectal forms, or if there was a lot of speech in overlap. These issues also affected annotation time for the correction. It took the second annotator approx. 90-105 minutes to correct 100 seconds of PCOMP annotations.

We introduced the same character (\uncertain) as an uncertainty marker as on the IPU layer. This process was also accompanied by regular meetings to discuss uncertainties, which contributed considerably to the development of segmentation guidelines and the label set.

\subsubsection{PCOMP validation}
\label{sec:PCOMPvalidation}

To validate the PCOMP annotations and to account for common confusions, we present a detailed evaluation of inter-annotator agreement of PCOMP annotations in this section.

To evaluate the quality of PCOMP annotations, we calculated Intra-Annotator agreement on 300 sec (100 sec from three conversations), labelled twice by one annotator and corrected by another, with some time in between the annotations.

In these 300 sec of conversation, 223 PCOMPs were annotated. 
The intra-annotator agreement for PCOMP boundaries was 0.96\footnote{Even though the first author did not annotate any of the other data, she also segmented the same 300 seconds for evaluation. The inter-annotator agreement for boundaries between this and the other two annotations by the primary annotator is 0.93 and 0.95.} (calculated for the boundaries that were set with respect to all the words where a boundary could have been set), indicating near-perfect agreement \citep[cf.,][]{LandisKoch1977}; .
In 13 out of 223 cases (6\%), a boundary was set in one annotation round but not in the other. In 7 of these cases, the resulting interval when a boundary was set was labelled as \disruption \hspace{0.5mm} or \incomplete. That is, in more than half of the cases of segmentation disagreements, speakers had not reached a point of potential completion. Two cases were annotated as \particle \hspace{0.5mm} and two cases as \cont, both of which are linked more closely syntactically with what follows than e.g., \hold \hspace{0.5mm} or \hes.

When comparing only the PCOMP labels in which the same boundaries were set, we obtained a Fleiss' $\kappa$ of 0.75 (based on 210 observations; z = 25.6, p $>$ .0001), indicating substantial agreement \citep{LandisKoch1977}. 
In 167 out of 210 observations, annotations were in agreement.

What these measures cannot capture, however, is when annotations were in partial agreement. In 14 out of these 43 observations, one or both PCOMPs were labelled with combined labels and one part of the combined label overlapped with the label in the other annotation round (e.g., \hrt \hspace{0.5mm} vs. {\fontfamily{cmtt}\selectfont hold\_hrt}, \hold \hspace{0.5mm} vs. {\fontfamily{cmtt}\selectfont hold\_question}, \hrt \hspace{0.5mm} vs. {\fontfamily{cmtt}\selectfont hrt\_question}, etc.). In the 29 cases in which annotations did not overlap at all, three confusions occurred more than once. First, \hrt \hspace{0.5mm} and \particle \hspace{0.5mm} (n = 3) were confused in or directly after overlap when it is not clear whether tokens such as “ja”, “naja” and “jaja” already take up the turn or not. Second, \particle \hspace{0.5mm} and \hold \hspace{0.5mm} (n = 4) were confused in overlapping speech. Third, \hold \hspace{0.5mm} and \cont \hspace{0.5mm} were confused in 11 cases. This was often due to different kinds of ellipses in the following PCOMP (e.g., ellipsis of the subject or of the main verb while the participle was expressed). The relatively frequent confusion of \cont \hspace{0.5mm} and \hold \hspace{0.5mm} in these cases indicates that a further demarcation of these two labels (e.g., in terms of different kinds of ellipses) is needed.
\begin{fp}{fp:029030Vaeter}{}
  \begin{center}
    \includegraphics[width=\linewidth]{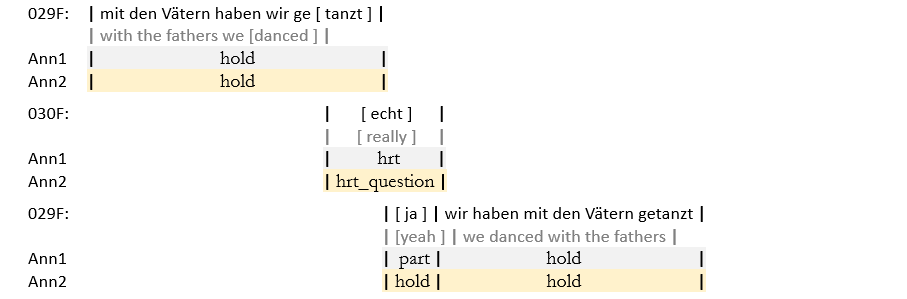}
    \label{Example_029030Vaeter}
  \end{center}
  \end{fp}
Sometimes, a whole sequence was analysed differently, which led to several disagreeing annotations in a row. In Example \ref{fp:029030Vaeter}, for instance, the annotators labelled the utterance “echt” (‘really’) by 030F as \hrt \hspace{0.5mm} in one annotation round and the subsequent turn-initial “ja” by 029F as \particle \hspace{0.5mm} (i.e., as an upbeat to the turn). In the other annotation round, they labelled “echt” as {\fontfamily{cmtt}\selectfont hrt\_question}, probably prompted by the strong, incredulous intonation with which it was uttered, and the subsequent turn-initial “ja” as \hold, because answers to questions are not labelled as \particle.

In conclusion, intra-rater agreement of our PCOMP turn-taking annotations is near perfect for boundary setting and substantial for PCOMP labels (cf., \cite{LandisKoch1977}). Most labelling disagreements are either only partial or can be explained by two different analyses of a sequence, both of which have their merits. It is, however, not possible to draw a comparison of this evaluation with other annotation systems.
None of the other studies investigating comparable phenomena \citep{LocalWalker2012, Zellers2017, zellers2019hand, Enomoto2020} have provided any measures of annotator agreement, since it is common practice in CA and CA-inspired studies to do a consensus annotation and exclude any uncertainties and disagreements from the analysis, which forestalls the need for such measures.

\section{A pilot study of conversational dynamics}
\label{sec:dynamics}

Most studies on turn-taking analyze stretches of conversation with respect to a particular research question, but not all continuous verbal productions of longer stretches of conversation. In this section, we aim to illustrate the added value of fully annotating longer stretches of speech by presenting how IPU and PCOMP annotations represent different conversational dynamics in individual conversations. 

\subsection{IPU dynamics}
\label{sec:dynamics_IPU}

\begin{figure}[ht]
    \centering
    \includegraphics[width=\linewidth]{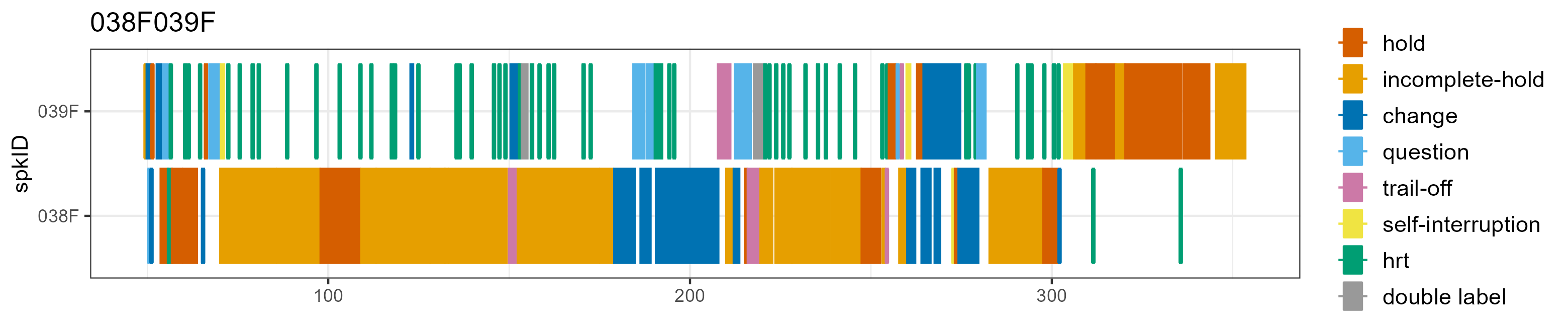}
    \caption{IPU annotations representing the dynamics of five minutes of conversation in 038F039F.}
    \label{fig:dynamics038039}
\end{figure}

\begin{figure}[ht]
    \centering
    \includegraphics[width=\linewidth]{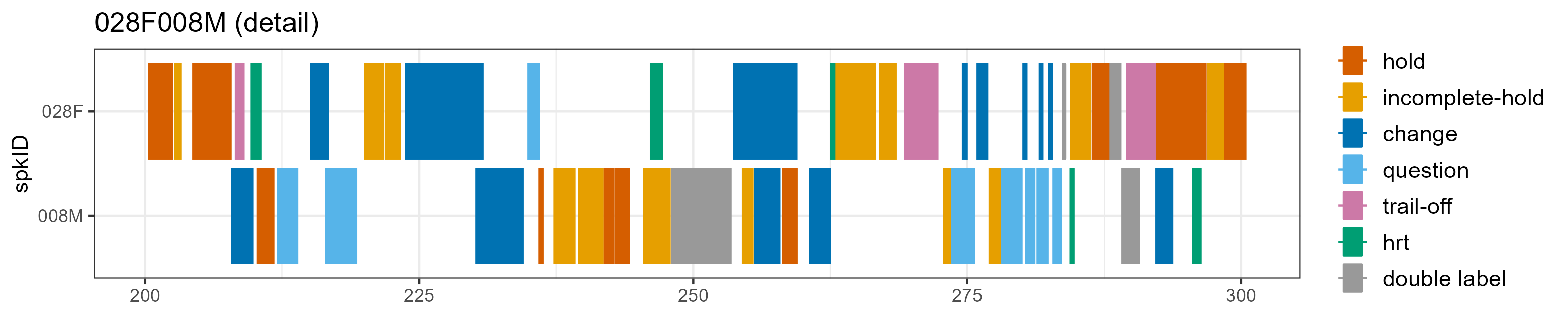}
    \caption{IPU annotations representing the dynamics of 100 seconds of conversation in 028F008M.}
    \label{fig:dynamics028008_IPU}
\end{figure}

\begin{figure}[ht]
    \centering
    \includegraphics[width=\linewidth]{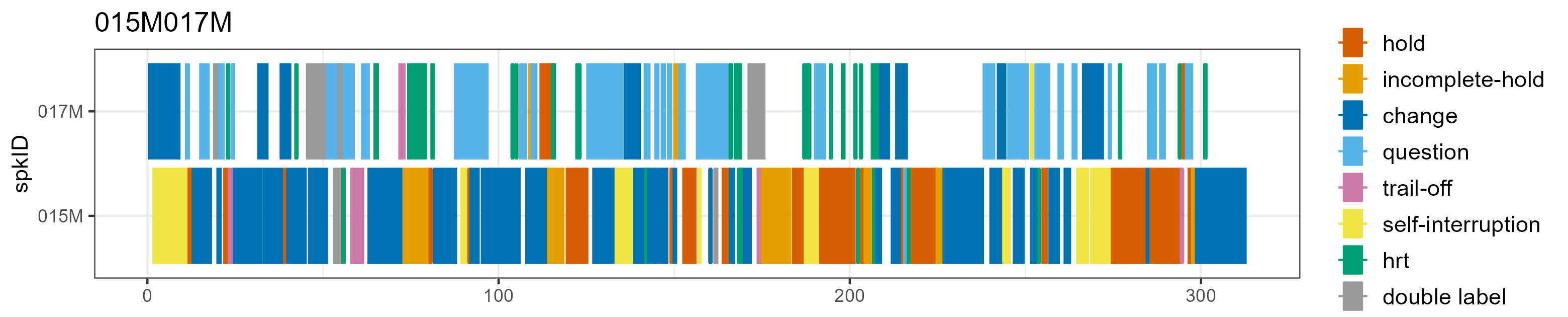}
    \caption{IPU annotations representing the dynamics of five minutes of conversation in 015M017M.}
    \label{fig:dynamics015017}
\end{figure}

IPU annotations capture different conversational dynamics. In the five minutes annotated in conversation 038F039F, for instance, 039F produced 53 {\hrt}s while her interlocutor 038F only produced three (cf., Table \ref{tab:IPUcounts}). 
On the other hand, 038F produced more, and generally very long, incomplete turn holds. Some of her IPUs are unusually long ($>$ 5 sec) because she tended to keep her turn by producing clearly audible inbreaths in her pauses. Accordingly, one long IPU was annotated instead of several short ones.
A temporal representation of IPU annotations in Figure \ref{fig:dynamics038039} further illustrates this asymmetric dynamic\footnote{To avoid too many colors in Figures \ref{fig:dynamics038039}, \ref{fig:dynamics028008_IPU}, \ref{fig:dynamics015017} and \ref{fig:dynamics028008_PCOMP}, which would make them unreadable, we cleaned the data before plotting. We grouped single uncertain labels with the respective certain single labels (e.g., {\fontfamily{cmtt}\selectfont hold@} with \hold, {\fontfamily{cmtt}\selectfont change@} with \change, etc.). We re-coded certain and uncertain combined labels as their macro-categories whenever possible (e.g. {\fontfamily{cmtt}\selectfont hold\_incomplete-hold} as a kind of hold, {\fontfamily{cmtt}\selectfont change\_question} as a kind of change, etc.; cf., \ref{sec:IPUvalidation}). Whenever the two parts of a combined label were not compatible in terms of their super-category (e.g., {\fontfamily{cmtt}\selectfont hold\_hrt}, {\fontfamily{cmtt}\selectfont change\_hrt}, {\fontfamily{cmtt}\selectfont hold\_change}, etc.), we grouped them into a residue class.}. 
038F mostly produced turn holds (red and orange), while 039F did not take up the turn for most of the annotated five minutes and almost exclusively produced {\hrt}s (green) and some {\qu}s (light blue). Only after 300 sec the dynamics changed and 039F took up a more active speaker role, while 038F took up the listener role, visible in the inverted proportion of turn holds (red and orange, 039F) and {\hrt}s (green, 038F). We observed that this asymmetry in turn-taking behaviour corresponds to different topics. At the beginning of the recording, 039F initiated the conversation by asking 038F about her day, which prompted a storytelling by 038F. 
At around 300 sec, the story-teller role shifted from one speaker to the other when 039F started to tell a related story of her own day. This narrative sequence by 038F that took up most of the snippet we are looking at here is also reflected in speaking time, which was twice as long (212 sec) compared to 039F (105 sec). While 038F’s speaking time was longer, she produced only about half the number of IPUs (41) compared to 039F (77), which is due to fewer but longer IPUs for 038F and more but shorter IPUs (mostly \hrt) for 039F.

An example for a more balanced conversation in terms of speaking time is a 5-minute stretch in conversation 028F008M. Table \ref{tab:IPUcounts} shows that these two speakers have a relatively similar distribution of annotation categories with the exception that 008M asked more questions (8 vs. 1 by 028F) and did not produce any self-interruptions or trail-offs (compared to very few by 028F; $<$ 5). Their speaking time is much more equally distributed (008M: 100 sec; 028F: 126 sec) than in 038F039F and they produced almost the same number of IPUs in these 5 minutes (008M: 62; 028F: 61). Both speakers produced a variety of different IPUs with a similar variation in durations. In terms of topic structure, this 5-minute sequence is characterised by several topic shifts, two of them disjunctive, a short narrative and by two short discussions.

A more detailed look at 100 seconds of this conversation in Figure \ref{fig:dynamics028008_IPU} can help us to identify shorter sequences of different conversational dynamics within this conversation. For instance, at around 235 sec, 028F asked a \qu \hspace{0.5mm} (light blue) which initiated a long turn by 008M which he realised as a number of IPUs annotated as syntactically complete and incomplete turn-holds (red and orange). Starting at around 270 sec, we find a sequence of {\qu}s by 008M that were followed by very short answers by 028F (cf., alternation of light blue and dark blue by the two speakers, respectively), until she took up a multi-unit turn (orange and red) after the fifth \qu.

Apart from speaking time and the duration of IPUs, another indication of conversational dynamics and of (a)symmetry in conversations is the distribution of annotation categories over speakers. In Figure \ref{fig:dynamics038039}, we already saw an example in conversation 038F039F in which one speaker almost exclusively produced \hrt, indicating that she mostly produced backchannels and continuers while the other speaker had the turn for most of the time.
The visual representation of five minutes in conversation 015M017M in Figure \ref{fig:dynamics015017}, for instance, illustrates that more than half of 017M’s speaking time was taken up by {\qu}s (light blue; almost half of his IPUs) and about a third of his IPUs were {\hrt}s (green), while half of 015M’s speaking time was composed of turn changes (dark blue; about one third of his IPUs). 
Most of the five minutes annotated in this conversation were taken up by 015M telling 017M about a boat that he had renovated, and 017M advanced this topic by asking many questions and produced many backchannels and continuers. While this conversation snippet is similar to 038F039F in that it is also characterised by one speaker having more epistemic authority over the topic, 017M took a more active role by asking a lot of questions about the topic, thereby also influencing the direction in which they were progressing the topic. In contrast, 039F (cf., Figure \ref{fig:dynamics038039}) almost exclusively produced \hrt, letting 038F decide in which direction to go in her story telling.

Another asymmetry in 015M017M is that 015M produced more self-interruptions (yellow) than 017M (9 vs. 1). This means that, when both speakers talked in overlap (e.g., when both started to talk simultaneously after a pause), 015M was the one who tended to interrupt himself to cede the turn to 017M. There is generally a lot of overlap in this conversation. However, since 017M produced so many questions, immediately transferring the turn back to 015M, we do not interpret this behaviour as turn-competition. 

In this section, we introduced a novel visual representation of the dynamics of stretches of conversations which is easy to extract from the annotations and is straightforward to read. It is useful at different stages of studies: for instance, in the phase of extracting data at the beginning of a study, in order to spot “visually” similar stretches of conversations across speaker pairs. Other uses are to find global differences of conversations across speaker pairs, or to study in detail how similar topics or activities are realized by different speaker pairs (e.g., narrative sequences in Figures \ref{fig:dynamics038039} and \ref{fig:dynamics015017} above).

\subsection{PCOMP dynamics}
\label{sec:dynamics_PCOMP}

\begin{figure}[ht]
    \centering
    \includegraphics[width=\linewidth]{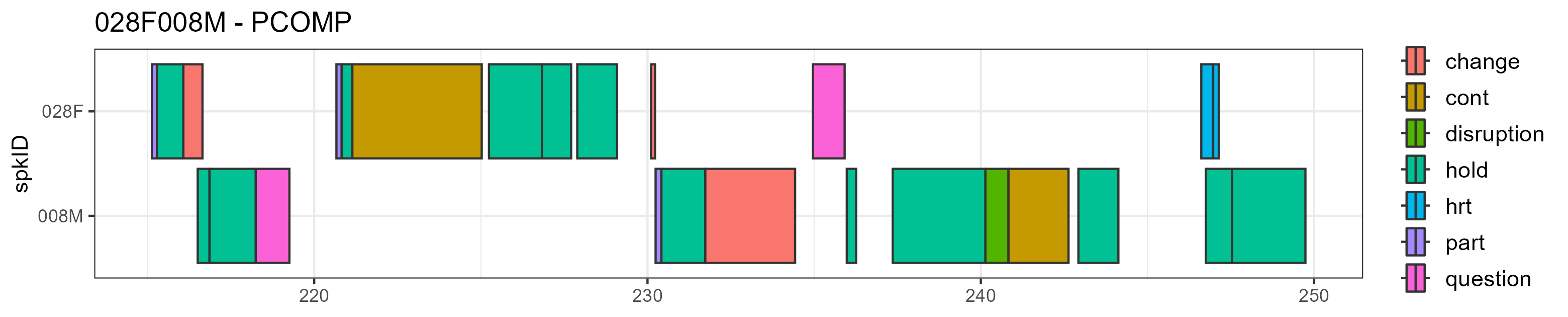}
    \caption{PCOMP annotations representing the dynamics and turn-internal structures in 35 seconds of conversation in 028F008M.}
    \label{fig:dynamics028008_PCOMP}
\end{figure}

\begin{figure}[h]
  \begin{adjustbox}{addcode={\begin{minipage}{\width}}{\caption{%
      Decision tree for assigning PCOMP labels, including single and frequent double labels. The first decision question is presented in the top left corner. For better readability branches are coloured in the same colours as in Figures \ref{fig:Flowchart_PCOMPsimple} and \ref{fig:dynamics028008_PCOMP}.
      }\end{minipage}},rotate=90,center}
      \includegraphics[scale=.639]{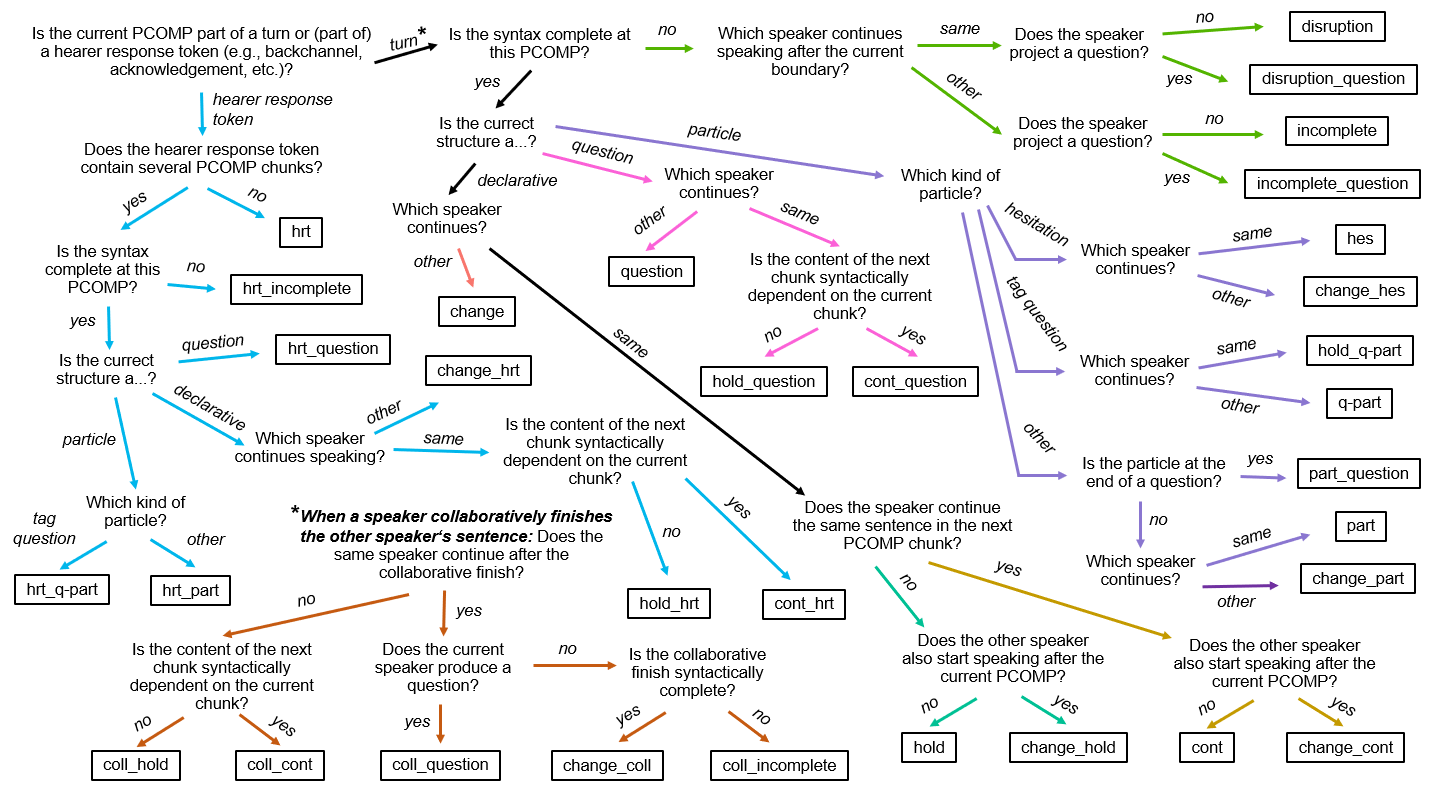}%
  \end{adjustbox}
  \label{fig:Flowchart_PCOMPcombined}
\end{figure}

Table \ref{tab:PCOMPcounts} in Section \ref{sec:PCOMPlabels} shows that the majority of turns consists of several TCUs. This is also illustrated in the snippet presented in Figure \ref{fig:dynamics028008_PCOMP}, which is a detail of the conversation shown in Figure \ref{fig:dynamics028008_IPU}. A colour-coded transcription of this snippet is presented in Example \ref{fp:PCOMP_dynamics} in \nameref{sec:appendixD}. In contrast to the more coarse IPU representation in Figure \ref{fig:dynamics028008_IPU}, the visual representation of PCOMP annotations here in Figure \ref{fig:dynamics028008_PCOMP} allows us to take a closer look at the internal structure of turns. Red and pink mark turn ends (\change{} and \qu, respectively), while different kinds of turn holds are coloured in the ocre-green-turquoise spectrum. Hearer response tokens that do not take up the turn are coloured blue.

Starting at 215 sec, 028F produced a turn consisting of a particle, a TCU labelled as \hold, and another discourse particle that was assigned the combined label {\fontfamily{cmtt}\selectfont change\_part} because it also constitutes the end of the turn (grouped here with \change{} for the visual presentation). In minimal overlap, 008M produced a turn consisting of two turn-holds and a question. After a gap, 028F produced a turn that included six PCOMPs and several turn-internal pauses. At 230 sec, 008M responded with a three-PCOMP turn. 028F responded with a simple question, which prompted a longer, multi-PCOMP turn by 008M with some longer pauses, which was only interrupted by a hearer response token.

This snippet includes several notable sequences that suggest several possible research questions. For instance, 028F ended the first turn in this snippet with a discourse particle. Discourse particles (except for tag questions) are rare in turn-final position. The label {\fontfamily{cmtt}\selectfont change\_part} could be filtered to investigate if turn-final discourse particles (in this case "halt") have different functions from the same particles in other, more common, sequential contexts.

Some turn-holds (\hold{} or \cont) were followed by a pause while others were directly followed by more speech. 
\cite{Sacks1974} argued that the unmarked case at a TRP is a speaker change. If a current speaker wants to hold the turn, they have to employ certain strategies to signal this intent. Various turn-holding strategies, such as level pitch accents in German \citep{selting1996}, or rush-through in English \citep{Walker2010} have been reported in the literature. \cite{kelterer2023}, however, did not find rush-through as a practice used in turn-holds before pauses in Austrian German. Thus, these annotations could be used to identify data for a study about whether speakers use different phonetic/prosodic turn-holding practices before pauses than before more speech without an intermediate pause.

This snippet includes two questions with a different make up. 008M's turn consists of several PCOMP intervals, while 028F's question only contains one PCOMP. She immediately threw back the ball into his court, even though she could have taken up more time by producing several PCOMPs leading up to her question, as 008M did. A comparison of question-answer sequences with single-PCOMP vs. multi-PCOMP questioning turns could address how they differ, for instance, in terms of topic progression, authority and engagement.

At around 244 sec, 008M made a long pause, after which he and 028F (producing a \hrt) started speaking simultaneously. Which prosody and which functions do hearer response tokens after such a long pause have in comparison to hearer response tokens timed immediately at another speaker's PCOMP? What do instances of a long within-turn pause plus a simultaneous start by both speakers have in common, for instance, in terms of prosody, topic structure, engagement or stance?

These examples illustrate how similar environments can be identified with these annotations, either by including timing criteria extracted from the time-aligned annotations in Praat, or by identifying them visually in figures like this one, and how they can suggest research questions concerning, for instance, turn design and various aspects of the communicative context.

\section{Conclusion}
\label{sec:conclusion}

This paper had two goals. First, we presented continuous turn-taking annotations in spontaneous dyadic conversations in GRASS in two domains that have frequently been used in the investigation of turn-taking; on the layer of Inter-Pausal Units (95 minutes annotated with near perfect agreement) and points of potential completion \citep[70 minutes annotated with substantial agreement; cf.,][]{LandisKoch1977}. This annotated corpus is available to the scientific community (cf., Section \ref{sec:availability}) and it is a resource suitable for qualitative and quantitative analyses, as well as for linguistic and technological investigations. 
Second, we presented the annotation system in more detail, including an analysis of fringe cases. This ensures replicability, so other researchers can apply this system to their own data from different domains, for instance, to conversations including power or knowledge hierarchies (e.g. doctor-patient, interviewer-interviewee), in language acquisition and autism research, or in the investigation of various kinds of speech pathologies.

The presented annotations have already been used in various studies. In Section \ref{sec:dynamics}, we presented an analysis of conversational dynamics.
\cite{kelterer2023} analysed the prosody of a turn-holding practice of pausing at a point of incomplete syntax.
\cite{ludusan2022analysis} investigated the role of laughter at turn boundaries.
Several topics are currently being investigated based on these annotations: the positioning and acoustics of breathings with respect to turn-taking \citep{Menrath}; listeners’ performance in transcription of speech with different kinds of disfluencies \citep[pause at syntactically incomplete point with continuation after the pause vs. with rephrasing after the pause vs. rephrasing without any preceding pause; ][]{Wepner_experiment}, and the prosody of hearer response tokens depending on their turn-taking context with the intention to implement it into a backchannel model for Furhat \citep{Paierl}.
Other examples of research questions that could be addressed with these turn-taking annotations are:
\begin{itemize}
  \item Which prosody do speakers display in self-interruptions when they continue speaking by rephrasing (\disruption), compared to self-interruptions in turn competition ({\fontfamily{cmtt}\selectfont self-\newline interruption})?

  \item Do speakers mark their turn-holds prosodically, depending on whether they continue the same syntactic structure with an increment ({\fontfamily{cmtt}\selectfont PCOMP-cont}) or start a new one in subsequent talk ({\fontfamily{cmtt}\selectfont PCOMP-hold})?
  \item Do speakers who produce a collaborative finish (\coll), that is, continue their interlocutor’s syntactic construction, also continue their interlocutor’s prosody?
  \item To which degree does \textit{prosodic} completion vs. continuation (cf., prosodic annotations in GRASS, Section \ref{sec:GRASS}) overlap with \textit{turn} completion (\change) vs. continuation (\hold, \cont)?
\end{itemize}

The presented corpus and the developed annotation system constitute a valuable resource for linguistics, speech science and technology and will hopefully, in the long term, inspire and facilitate the investigation of turn-taking across disciplines.

\section{Corpus availability}
\label{sec:availability}
The corpus, i.e., the audio along with the IPU and PCOMP annotations in Praat Textgrid format, is freely available for non-commercial research. 
More information on GRASS can be found on 
\url{https://www.spsc.tugraz.at/databases-and-tools/grass-the-graz-}\newline\url{corpus-of-read-and-spontaneous-speech.html}.

\section{Acknowledgements}
This research was funded by the Austrian Science Fund (FWF) [10.55776/P32700]. We thank Sophie Christian, Dorina Gabler and Jakob Murauer for their efforts in creating the turn-taking annotations for the GRASS corpus and for fruitful discussions during the annotation process.

\section*{Appendix}
\label{sec:appendix}
\newpage
\subsection*{Appendix A}
\label{sec:appendixA}

\begin{table}[ht]
\centering
\caption{Conversation and time stamp of examples from GRASS presented in this paper.}
\vspace{+10pt}
\begin{tabular}{lll}
  \hline 
Example & Conversation & Time stamp \\ 
  \hline  
  Figure \ref{fig:Praat} &  003M023F & 00:02:40 \\
  Example \ref{fp:004M024Fspektatulär} &  004M024F & 00:04:02 \\ 
  Figure \ref{fig:Praat_IAA} &  029F030F & 00:00:53 \\
  Example \ref{fp:meine004024} &   024F004M  & 00:03:37 \\  
  Example \ref{fp:irgendwas001002} &  001M002M  & 00:00:36 \\ 
  Example \ref{fp:013014auskennst} &  013M014M  & 00:00:24 \\ 
  Example \ref{fp:zuFuss} &   003M023F   & 00:00:36 \\ 
  Example \ref{fp:Platine} &   006M007M  & 00:02:27 \\ 
  Example \ref{fp:Keplerbruecke} &  003M023F  & 00:00:43 \\ 
  Example \ref{fp:Boot} &   015M017M & 00:02:24 \\  
  Example \ref{fp:zufrieden038039} &   038F039F  & 00:01:07 \\ 
  Example \ref{fp:wohnen001002} &  001M002M  & 00:04:53 \\ 
  Example \ref{fp:029030Vaeter} &  029F030F & 00:02:28 \\  
  Example \ref{fp:PCOMP_dynamics} &  028F008M  & 00:03:35 \\  
   \hline 
\end{tabular}
\label{tab:timestamps}
\end{table}

\clearpage

\subsection*{Appendix B}
\label{sec:appendixB}

\clearpage

\subsection*{Appendix C}
\label{sec:appendixC}

\clearpage

\subsection*{Appendix D}
\label{sec:appendixD}

Transcript of the PCOMP annotations presented in Figure \ref{fig:dynamics028008_PCOMP}.

\begin{fp}{fp:PCOMP_dynamics}{}
\begin{center}
  \includegraphics[width=\linewidth]{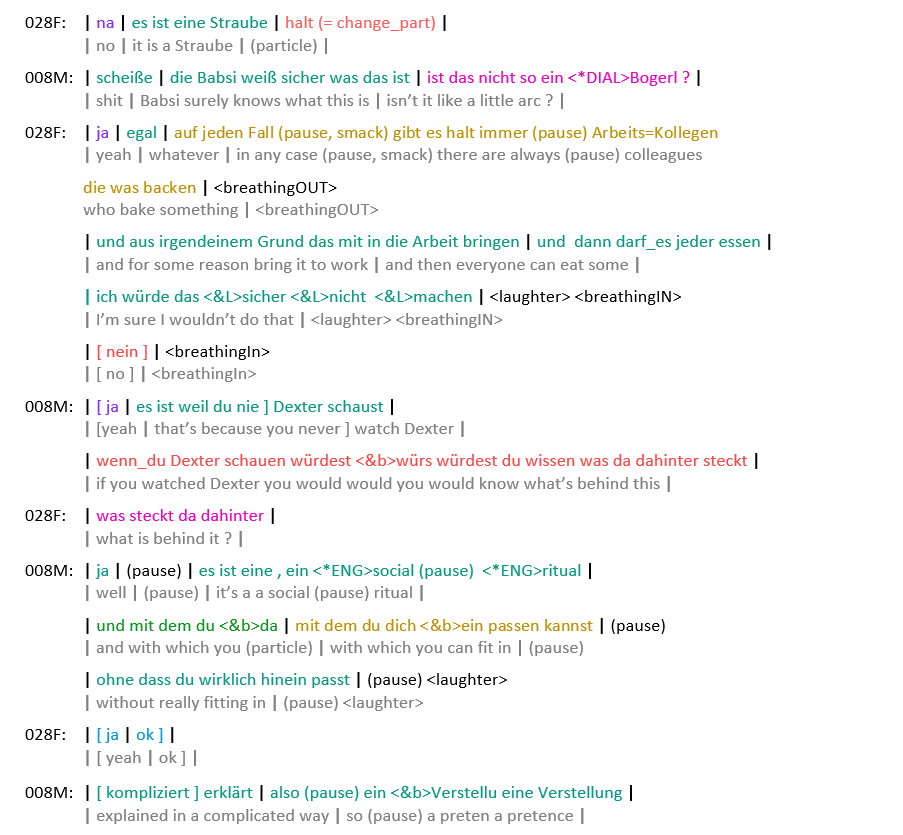}
  \label{Example_PCOMP_dynamics}

\end{center}
\end{fp}

\newpage
\bibliography{Annotation_paper}

\end{document}